\documentclass[11pt,table]{article}
\usepackage{amsmath}
\usepackage[utf8]{inputenc}
\usepackage{textgreek}
\usepackage{amssymb}

\usepackage[preprint]{acl}
\usepackage{subcaption}
\usepackage{multirow}
\usepackage{pifont}

\usepackage[dvipsnames,table]{xcolor}
\usepackage[most]{tcolorbox}
\usepackage{graphicx}
\usepackage{float}
\usepackage{booktabs}
\usepackage{tabularx}
\usepackage{array}
\usepackage{cuted}
\usepackage{listings}
\usepackage{mdframed}

\tcbuselibrary{breakable,listings}

\definecolor{DarkGreen}{RGB}{0,100,0}

\usepackage{algorithm}
\usepackage{algpseudocode}
\algrenewcommand\algorithmicrequire{\textbf{Input:}}
\algrenewcommand\algorithmicensure{\textbf{Output:}}
\algrenewcommand{\algorithmiccomment}[1]{\hfill$\triangleright$ #1}

\newcounter{myboxcounter}

\newtcolorbox{mybox}[2][]{
    before upper={
        \refstepcounter{myboxcounter}
    },
    colback=cyan!3,
    colframe=cyan!25!blue!75,
    title=\textbf{#2},
    breakable,
    #1
}

\definecolor{A}{RGB}{255,248,220} 
\definecolor{B}{RGB}{255,235,170} 
\definecolor{avggray}{RGB}{235,235,235}
\definecolor{green}{RGB}{190,225,190}

\definecolor{rqheader}{HTML}{1b8dff}    
\definecolor{rqbg}{HTML}{d2e9ff}       

\definecolor{promptheader}{HTML}{ff5d5a}      
\definecolor{promptbg}{HTML}{fff1f1}   
\definecolor{border}{RGB}{120,140,180}    

\newmdenv[
  topline=false,
  bottomline=false,
  rightline=false,
  leftline=true,
  linecolor=border,
  linewidth=3pt,
  innertopmargin=4pt,
  innerbottommargin=4pt,
  innerleftmargin=10pt,
  innerrightmargin=4pt,
  skipabove=4pt,
  skipbelow=4pt
]{algobox}

\usepackage{times}
\usepackage{latexsym}

\usepackage[T1]{fontenc}

\usepackage[utf8]{inputenc}

\usepackage{microtype}

\usepackage{inconsolata}

\definecolor{PreprocessBlue}{HTML}{1565C0}
\definecolor{ChunkOrange}{HTML}{EF6C00}
\definecolor{COPEGreen}{HTML}{2E7D32}
\definecolor{IntentPurple}{HTML}{6A1B9A}
\definecolor{RetrieveRed}{HTML}{C62828}
\definecolor{GenerateTeal}{HTML}{00897B}
\definecolor{VerifyCyan}{HTML}{00838F}

\title{LabourCrew: A Multi-Agent RAG Framework for Trustworthy Adversarial Deliberation and Statutory Reasoning over Labour Law}

\author{
 \textbf{Fatema Tuj Johora Faria\textsuperscript{1}},
 \textbf{Mukaffi Bin Moin\textsuperscript{1}},
 \textbf{Jubayer Al Mahmud\textsuperscript{2}},\\
 \textbf{M. F. Mridha\textsuperscript{3}},
 \textbf{Md. Alam Hossain\textsuperscript{2}}
\\
\\
 \textsuperscript{1}Ahsanullah University of Science and Technology, Bangladesh\\
 \textsuperscript{2}Jashore University of Science and Technology, Bangladesh\\
 \textsuperscript{3}American International University - Bangladesh\\
 \small{
   \textbf{Correspondence:} \href{mailto:mukaffi28@gmail.com}{mukaffi28@gmail.com}, \href{mailto:fatema.faria142@gmail.com}{fatema.faria142@gmail.com} 
 }
}

\begin{document}
\maketitle

\begin{abstract}

In statutory question answering, every claim must be traceable to evidence, not merely relevant, since unverifiable labour-rights answers carry serious legal consequences. Current systems fall short: single-pass RAG cannot detect insufficient evidence, while multi-agent legal-debate systems treat grounding as a prompting convention, letting agents cite unretrieved evidence. To address this gap, we introduce \textbf{LabourCrew}, a multi-agent RAG framework built around three grounding mechanisms: \textbf{StatuteGraph}, a graph index that explicitly links chapter, section, proviso, and cross-reference structure rather than fixed-length spans; an \textbf{Evidence Exchange Protocol} that confines advocates and an interpreter to an evidence ledger, making citation to unretrieved text impossible, while a fault-tolerant supervisor board runs advocates in parallel so individual failures degrade rather than crash the system; and a \textbf{Calibrated Trust Gate} that replaces categorical accept/reject decisions with a trust score, thresholded via conformal risk control for a distribution-free bound on the false-accept rate. We evaluate on \textbf{LabourActQA}, a 500-item Bangla question set from the Bangladesh Labour Act, 2006, spanning seven reasoning categories and three difficulty tiers. The framework drives the empirical false-accept rate to 0.081, within the target level ($\alpha = 0.10$), achieves the highest Answer Relevancy among HyDE RAG, Graph-RAG, and Hierarchical RAG (0.862 $>$ 0.839, 0.815, 0.828), and degrades gradually rather than catastrophically as question difficulty increases. These results show that calibrated abstention, not retrieval quality alone, is what makes legal question answering auditable in low-resource statutory domains.

\end{abstract}

\section{Introduction}
\label{sec:1}

\begin{figure}[h]
    \centering
    \includegraphics[width=\linewidth]{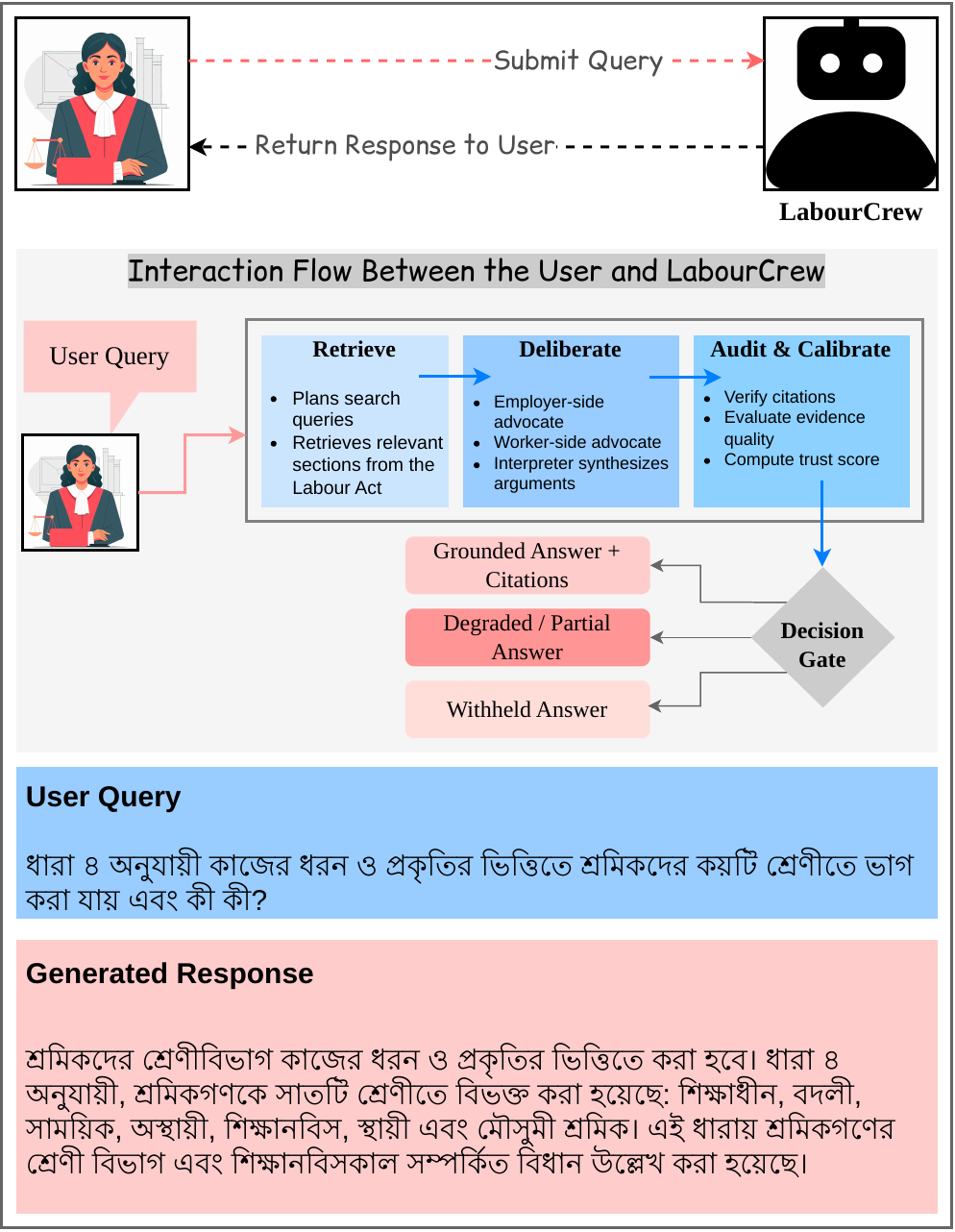}
    \caption{How \textbf{LabourCrew} answers a query: \textbf{Retrieve} $\rightarrow$ \textbf{Deliberate} $\rightarrow$ \textbf{Audit \& Calibrate} $\rightarrow$ a grounded, degraded, or withheld answer.}
    \label{intro-example}
\end{figure}

Legal NLP systems support decisions where wrong answers carry material consequences; labour law is canonical, since an erroneous notice period or wage entitlement can affect a worker's income.\footnote{We use the British spelling \textit{labour} throughout, following the official name of the statute under study, the Bangladesh Labour Act, 2006.} Current LLM-plus-retrieval pipelines \citep{lewis2020rag} are poorly matched to statutory text: fixed-size chunking separates a section from its qualifying proviso, so accurate search still returns an incomplete unit a static pipeline cannot correct. This is not hypothetical: hallucination remains an open problem \citep{huang2023hallucination}, legal-RAG audits show systems marketed as reliable hallucinate at 17--33\% \citet{magesh2024}, and surface correctness does not imply a claim was derived from its citation \citep{wu2024}.

Multi-agent approaches move legal reasoning beyond single-pass generation: judge/prosecutor/lawyer simulation improves judicial decisions \cite{he2024}; adversarial lawyer roles improve courtroom debate \cite{chen2025agentcourt}; multi-agent collaboration surfaces legal-theory insights a single chain misses \cite{yuan2024}, building on findings that debate improves factual accuracy \cite{du2024}. None specify who may access evidence; citation is left to instruction, not the tool level — so grounding remains violable and none recovers from failure.

A parallel line addresses evaluation and retrieval rather than grounding: one benchmark formalizes multi-agent legal reasoning over GDPR compliance, favoring specialization over end-task accuracy \cite{jing2025}; a survey of legal agents names hallucination and evaluation methodology as unresolved \cite{liu2026}. Graph-structured \cite{han2025} and on-demand retrieval \cite{asai2024} are studied outside law; the closest neighbor applies graph retrieval to law via a hierarchical knowledge graph, but sequentially, with no parallel deliberation or fault isolation \cite{chen2026legalgraphrag}.

Statistical calibration attaches a distribution-free error guarantee to an LLM's output rather than an unvalidated cutoff: conformal factuality calibration bounds claim-selection error in single-pass generation \cite{mohri2024}, building on conformal risk control's threshold-against-held-out-data risk bound \cite{angelopoulos2022,angelopoulos2021}. This calibrates only raw model confidence so far, never a trust signal from audited deliberation, nor in law. No system offers structure-aware retrieval, protocol-enforced grounding, fault isolation, and calibrated release together — a gap acute outside high-resource jurisdictions \cite{jing2025,liu2026}.

We present \textbf{LabourCrew}, an evidence-gated multi-agent RAG framework for statutory reasoning over the Bangladesh Labour Act. The statute is indexed by \textbf{StatuteGraph}, a structure-preserving parser reconstructing chapter, section, proviso, and cross-reference as a linked graph rather than fixed-length windows; an \textbf{Issue Spotter}, \textbf{Retrieval Planner}, \textbf{Link Hopper}, and \textbf{Statute Retriever} decide what to query and expand along the statute's own links in an interleaved reason-then-act loop \citep{yao2023react}. Retrieved evidence is deliberated by a \textbf{Worker Counsel} and \textbf{Employer Counsel} arguing opposing readings, and a \textbf{Legal Interpreter} applying the governing rule; all three communicate only through the \textbf{Evidence Exchange Protocol (EEP)}, citing only node IDs already on the shared ledger, since the debate agents have no retrieval tool of their own.

A \textbf{Trust Auditor} monitors deliberation for missing evidence and contradictions, while a \textbf{fault-tolerant Supervisor} runs advocates in parallel and isolates any single role's failure, so an agent's error degrades rather than collapses the pipeline. A \textbf{Citation Checker} mechanically verifies every cited node and span before a release-time \textbf{Trust Gate}, calibrated via conformal risk control \cite{mohri2024,angelopoulos2022}, decides release as a \textbf{Trustworthy Grounded Legal Opinion (TGLO)}, or marks it degraded or undecided, with a distribution-free error guarantee. We evaluate on \textbf{LabourActQA}, a 500-item, seven-category set over the Act spanning three difficulty tiers, under-served relative to prior legal NLP's high-resource corpora.

This architecture lets us test five research questions isolating each design choice's contribution:

\begin{itemize}
\item \textbf{RQ1:} Does structure-preserving retrieval yield better grounding and efficiency than flat, similarity-only retrieval?
\item \textbf{RQ2:} Does the Evidence Exchange Protocol's tool-level restriction reduce unsupported claims versus unrestricted multi-agent debate?
\item \textbf{RQ3:} Does fault-tolerant, supervised orchestration reduce failed runs and recover from failure, versus an unsupervised sequential pipeline?
\item \textbf{RQ4:} Is the calibrated Trust Gate more effective than an uncalibrated, categorical decision, and does its guarantee hold empirically?
\item \textbf{RQ5:} Does performance degrade gracefully with task difficulty, or does reliability collapse on the hardest categories despite strong aggregate scores?
\end{itemize}

\section{Related Work}
\label{sec:2}

\subsection{Faithfulness and Trust Calibration in Retrieval-Augmented Legal Generation}
\label{sec:2-1}

Recent work asks whether retrieval-augmented generation makes legal LLM output trustworthy, not merely plausible. Audits of commercial legal-RAG tools find hallucination reduced but not eliminated, error rates still 17--33\% \cite{magesh2024}. A complementary strand interrogates the retrieval-generation link itself: correctly citing a passage does not imply the claim was derived from it, since models often post-rationalize by attaching a citation to an already-formed assertion \cite{wu2024}. A separate thread bounds this error with finite-sample guarantees: conformal factuality calibration wraps single-pass generation with a claim-selection procedure provably controlling error rate \cite{mohri2024}, building on conformal risk control's expected-risk bound over held-out data \cite{angelopoulos2022}, with a PAC-style alternative \cite{angelopoulos2021}; selective conformal procedures test whether a sample departs from the calibration distribution before admitting it \cite{wang2025sconu}, but still calibrate raw model output, not audited deliberation. Audits are diagnostic, not corrective; calibration supplies the missing mechanism but has so far bounded only raw confidence, never in the legal domain.

\subsection{Multi-Agent Legal Reasoning and Debate}
\label{sec:2-2}

Beyond single-pass generation, several systems adopt role-differentiated, multi-agent legal reasoning mirroring legal process: court-simulation with judge, prosecutor, and lawyer roles improves judicial decision-making \cite{he2024}; adversarial, evolvable lawyer roles improve courtroom debate across repeated trials \cite{chen2025agentcourt}; multi-agent collaboration surfaces legal-theory insights a single chain misses, improving charge prediction \cite{yuan2024}. Decomposing reasoning across roles improves quality and realism over a monolithic LLM, but grounding is enforced only as a prompting convention — citation isn't restricted at the tool level — and none report a failure-recovery mechanism, targeting case-law prediction in high-resource jurisdictions, not statutory interpretation.

\subsection{Structured Retrieval and Benchmarking for Legal Multi-Agent Systems}
\label{sec:2-3}

Other work adapts structured, agentic retrieval and evaluation protocols to legal multi-agent pipelines, part of a broader shift from static retrieval toward agents that plan, reflect, and adapt what they retrieve \cite{singh2025agenticrag}. MASLegalBench, the first multi-agent deductive legal reasoning benchmark, uses GDPR compliance questions to argue for evaluating specialization over end-task accuracy alone \cite{jing2025}. A recent survey names hallucination, verifiability, and evaluation methodology as legal agents' central unresolved challenges \cite{liu2026}. Closest to this work, LegalGraphRAG replaces flat retrieval with a hierarchical knowledge graph, routed sequentially through Researcher $\to$ Auditor $\to$ Adjudicator, for judgment prediction \cite{chen2026legalgraphrag}; L-MARS decomposes a query across agentic web-search and a citation-faithfulness verification agent, but over the open web, without a calibrated release gate \cite{wang2025lmars}. Limitations are specific: MASLegalBench targets multiple-choice compliance in one high-resource regime, not open-form statutory QA; the survey names challenges without operationalizing a system; LegalGraphRAG is sequential, with no parallel roles or fault isolation, and a heavyweight graph rather than a lightweight statute index; L-MARS verifies against the open web, not a versioned graph, with no statistical release bound.

\subsection{Research Gap and Positioning of the Proposed Work}
\label{sec:2-4}

These three strands (Sections~\ref{sec:2-1}--\ref{sec:2-3}) leave a common gap unaddressed: no existing system unifies statute-aware retrieval, protocol-enforced grounding, fault-isolated multi-agent reasoning, and a statistically calibrated release decision within a single pipeline. LabourCrew is positioned at this intersection, applied to a low-resource statutory jurisdiction, the Bangladesh Labour Act.

\section{LabourCrew Framework}
\label{sec:3}

\subsection{Problem Definition}
\label{sec:3-1}

\begin{figure*}[h]
\centering
\includegraphics[width=\textwidth]{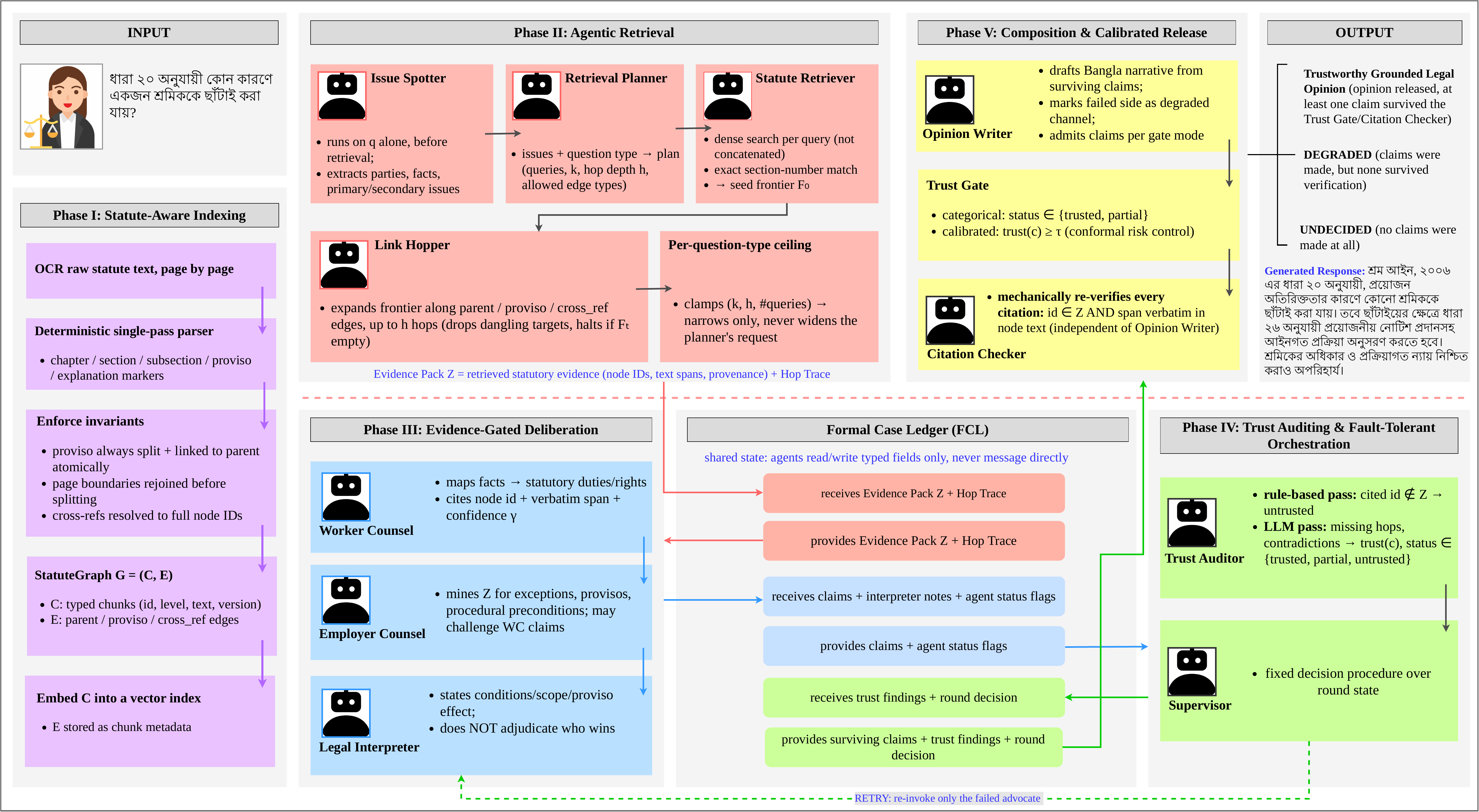}
\caption{The \textbf{LabourCrew} architecture: Phase I builds \textbf{StatuteGraph} offline; Phases II--V run per query, retrieving evidence, deliberating over it, auditing and retrying failed roles, then composing and gating release as \textbf{TGLO}, \textbf{DEGRADED}, or \textbf{UNDECIDED}. All agents communicate only through the shared \textbf{Formal Case Ledger (FCL)}.}
\label{fig:LabourCrew_framework}
\end{figure*}

\textbf{Task.} Given a natural-language legal question $q$ about a labour statute, the system must produce a \textbf{Trustworthy Grounded Legal Opinion (TGLO)} $\hat{y}$: a narrative answer in which every substantive claim is traceable to a verbatim, existing statutory passage, with an explicit release status $s \in \{\text{TGLO}, \text{DEGRADED}, \text{UNDECIDED}\}$.

\textbf{Statute representation.} The statute is a graph $G = (C, E)$, where $C = \{c_1, \dots, c_n\}$ is a set of legal chunks and $E \subseteq C \times C \times \{\texttt{parent}, \texttt{proviso}, \texttt{cross\_ref}\}$ is a set of typed directed edges, built once offline and fixed for all questions.

\textbf{Claims and evidence.} A claim $c = (\text{side}, \text{text}, R, \gamma)$ consists of an authoring side $\text{side} \in \{\text{Worker}, \text{Employer}\}$, an assertion, evidence references $R = \{(\text{id}, \text{span})\}$, and a self-reported confidence $\gamma \in [0, 1]$.

\textbf{Assumptions.} $G$ is assumed complete and correctly versioned at query time; claims are produced only by agents with access to the retrieved evidence pack (Section~\ref{sec:3-5}), and no agent may introduce evidence absent from $G$.

\label{sec:3-2}
\textbf{LabourCrew} is a directed graph of eleven role-specialized agents operating over a single shared state object, the \textbf{Formal Case Ledger (FCL)}, rather than a linear pipeline: agents communicate only through the \textbf{Evidence Exchange Protocol} (EEP), reading and writing typed fields on the FCL, and a supervising controller mediates every phase transition, letting retrieval react to what deliberation later finds missing without a single component's failure silently propagating. The only loop-backs are retrying a failed advocate and requesting further retrieval; the final Trust Gate fails closed on a rejected claim, guaranteeing termination.

\subsection{Phase I: Statute-Aware Indexing (StatuteGraph)}
\label{sec:3-3}

This phase constructs $G$ from raw statute text via a deterministic, single-pass parser, run once offline rather than by any agent, since fixed-length and recursive-character chunking routinely separate a rule from the proviso that qualifies it, leaving a retrieved chunk legally incomplete even when topically correct. It preserves chapter, section, proviso, and cross-reference structure as typed edges, so every downstream trust mechanism can verify a claim against the chunk it cites. The parser walks the OCR-extracted text block by block, tracking the currently open chapter, section, and subsection, and classifies each block against the statute's own typographic markers (chapter headings, numbered sections including amendment-lettered ones, subsection numbers, proviso and explanation markers, lettered sub-clauses) rather than by position or length, so a chunk boundary only falls where the statute's own numbering says a new unit begins. A proviso always splits into its own chunk, with the parent-child link written atomically so it cannot exist without a parent reference; OCR page boundaries are rejoined before parsing, using clause-final punctuation to judge whether a unit continues onto the next page; and a cross-reference naming several sections in one list resolves to every named section, not only the first. The resulting chunks are tagged with a statute-version identifier and embedded into a vector index, while the edge set is stored as chunk metadata rather than a separate graph database.

\subsection{Phase II: Agentic Retrieval}
\label{sec:3-4}

Given $q$, this phase retrieves an \textbf{Evidence Pack} $Z \subseteq C$ covering the operative rule and any qualifying proviso or cross-reference, deciding on demand how much structure to traverse rather than retrieving once. Three agents act here (\textbf{Retrieval Planner}, \textbf{Statute Retriever}, \textbf{Link Hopper}), consuming the issues and question-type label the \textbf{Issue Spotter} derived from $q$ alone, together with, on re-entry, a structured retrieval request from the Trust Auditor naming seed nodes, edge types, and a hop budget. The Retrieval Planner converts these into a plan: a set of semantic queries, a target dense-hit count $k$, a maximum hop depth $h$, and the allowed edge types to traverse. The Statute Retriever then runs each semantic query independently, since a single joined query loses recall on questions spanning two concepts, unioning the top-$k$ hits into a seed frontier, while section numbers named directly in $q$ are resolved separately by exact suffix match. The Link Hopper expands that frontier for $h$ hops along the plan's allowed edge types, fetching each hop's unvisited targets and halting early once a hop returns nothing new, so a governing proviso ranking low on similarity is still recovered structurally; the Statute Retriever finally fuses the dense and hopped node sets, deduplicated by identifier, with the executed query count, $k$, and $h$ all clamped to a fixed per-question-type ceiling that can only narrow, never widen, a plan. This phase also writes a Hop Trace, recording the source node, edge type, and hop depth for every expanded node, to the FCL, letting the Trust Auditor later check \textit{why} a node is present, not merely whether.

\subsection{Phase III: Evidence-Gated Deliberation}
\label{sec:3-5}

This phase produces claims and counter-claims arguing opposing readings of the retrieved evidence, and a neutral statement of the governing rule, via four agents: the \textbf{Issue Spotter}, which intakes $q$ before retrieval begins and extracts the case seeds (parties, facts, primary and secondary issues) that both this phase and Phase II (Section~\ref{sec:3-4}) depend on; and the \textbf{Worker Counsel}, \textbf{Employer Counsel}, and \textbf{Legal Interpreter}, who deliberate once $Z$ is available. Grounding is enforced by restriction, not instruction: the advocates have no tool to retrieve outside $Z$, so a claim citing outside evidence is not constructible. Their roles are asymmetric: the Worker Counsel maps case facts to statutory duties owed, while the Employer Counsel mines $Z$ for exceptions and preconditions that defeat those duties and may directly challenge a Worker Counsel claim; neither depends on the other, so a failed call is recorded per-agent rather than aborting the round, and on later rounds each side also sees the opposing side's existing claims. The Legal Interpreter then runs once over whichever claims are present, one-sided if one advocate failed, applying the operative rule without declaring which side prevails; the phase closes by writing the resulting claim set, tagged \texttt{ok}, \texttt{partial}, or \texttt{failed} at the agent level, with the Interpreter's notes, to the FCL.

\subsection{Phase IV: Trust Auditing and Fault-Tolerant Orchestration}
\label{sec:3-6}

This phase determines, each round, whether claims are adequately grounded and whether to compose, retry a failed role, retrieve further, or abort gracefully. The \textbf{Trust Auditor} computes a deterministic composite trust score for every claim $c$:
\begin{equation}
\small
\begin{aligned}
\operatorname{trust}(c) ={}&
w_v \operatorname{validity}(c)
+ w_f \operatorname{fidelity}(c) \\
&+ w_p \operatorname{provenance}(c)
+ w_s \gamma(c),
\end{aligned}
\end{equation}
with $w_v = 0.35$, $w_f = 0.25$, $w_p = 0.20$, $w_s = 0.20$ (summing to 1): $\text{validity}(c)$ is the fraction of cited identifiers that exist in $Z$, $\text{fidelity}(c)$ is the fraction of cited spans found verbatim (whitespace-normalized) in their node's text, with a missing or empty span counted as unverified; $\text{provenance}(c)$ is the mean retrieval provenance over valid citations; and $\gamma(c)$ is the advocate's own confidence, clipped to $[0,1]$; a claim with no evidence receives $\text{trust}(c)=0$ unconditionally. This score is computed identically regardless of what produces the accompanying categorical finding: a deterministic rule-based pass first marks any claim whose cited identifiers do not all exist in $Z$ as \texttt{untrusted}, after which an LLM pass audits the remaining claims for missing required hops and inter-side contradictions, producing a status of \texttt{trusted}, \texttt{partial}, or \texttt{untrusted} and, where applicable, a structured retrieval request; any claim for which the LLM pass returns no finding is assigned \texttt{untrusted} explicitly. The \textbf{Supervisor} then applies a fixed decision procedure over the round state: at round zero, if exactly one advocate failed, it retries only that advocate; otherwise, if an untrusted finding carries a retrieval request and the round budget is not exhausted, it requests retrieval; otherwise, if the budget is exhausted or at least one finding is not \texttt{untrusted}, it proceeds to composition; and if the budget is nearly exhausted with no trusted evidence, it aborts softly to composition with whatever exists. Isolating advocate failures per-agent and retrying only the failed role bounds the cost of a transient failure to the one component that actually failed.

\subsection{Phase V: Composition and Calibrated Release Gate}
\label{sec:3-7}

This final phase drafts an opinion from surviving claims and decides --- with a statistically bounded error rate rather than an arbitrary cutoff --- whether to release, partially release, or withhold it. The \textbf{Opinion Writer} drafts the narrative from whichever claims survived, explicitly marking any side whose advocate failed as a degraded channel, and produces a set of accepted and rejected claim identifiers; quoted evidence spans are never altered in translation, since the Citation Checker's verification requires them byte-identical, modulo whitespace, to the source statute text. Claim admission is then governed by the active gate mode: under \texttt{categorical} mode, a claim is accepted iff its trust finding's status is \texttt{trusted} or \texttt{partial}; under \texttt{calibrated} mode, a claim is accepted iff $\text{trust}(c) \geq \tau$, where $\tau$ is set offline by \textbf{Conformal Risk Control}: given $n$ labeled calibration pairs and target risk $\alpha$, the smallest $\tau$ satisfying
\begin{equation}
\frac{n}{n+1}\,\hat{R}(\tau) + \frac{1}{n+1} \;\leq\; \alpha,
\end{equation}
where $\hat{R}(\tau)$ is the empirical false-accept rate, guarantees $\mathbb{E}[\text{false-accept rate}(\tau)] \leq \alpha$ over future exchangeable claims, replacing a categorical accept/reject decision with a provable bound; if no threshold in the candidate grid satisfies the bound, the gate fails closed ($\tau = +\infty$, rejecting every claim). Independently of gate mode, the \textbf{Citation Checker} then mechanically re-verifies every citation the accepted claims make, confirming the cited identifier exists in $Z$ and its span is found verbatim, whitespace-normalized, in the cited node's text, stripping any claim that fails this check from the accepted set rather than looping the whole board again; the Citation Checker and the trust score's fidelity term share one whitespace-normalization routine, so a genuinely correct quotation cannot pass the gate and then be silently stripped, or vice versa, purely over OCR formatting. The release status then follows directly from what remains: $\texttt{TGLO}$ if at least one claim is accepted, $\texttt{DEGRADED}$ if claims were made but none survived, and $\texttt{UNDECIDED}$ if none were made, alongside a validation report of citations checked, passed, and failed, and a Trustworthiness Scorecard covering evidence coverage, path completeness, and citation precision.

\section{Dataset Construction}
\label{sec:4}

Prior legal-NLP benchmarks target contract review \citep{hendrycks2021cuad}, English-language legal understanding \citep{chalkidis2022lexglue}, general legal reasoning \citep{guha2023legalbench}, or civil-code statute retrieval and entailment in a different jurisdiction \citep{goebel2023coliee}; low-resource Bangla NLP resources, in turn, target generation and understanding tasks outside the legal domain \citep{bhattacharjee2023banglanlg}. No existing benchmark evaluates statutory reasoning over Bangla labour law, so we construct \textbf{LabourActQA}, a 500-item question-answering resource built directly over the Bangladesh Labour Act, 2006, used exclusively for evaluation. Full construction protocol, annotation schema, and quality-control detail are in Appendix~\ref{sec:A}.

\begin{table*}[h]
\centering
\scriptsize
\begin{tabular}{lrrrrrr}
\toprule
\rowcolor{rqbg}
Method / Configuration & N & Answer Relevancy $\uparrow$ & Context Relevancy $\uparrow$ & Context Precision $\uparrow$ & Context Recall $\uparrow$ & Noise Sensitivity $\downarrow$ \\
\midrule
\textbf{LabourCrew} & \textbf{500} & \textbf{0.862} & \textbf{0.891} & \textbf{0.914} & \textbf{0.913} & \textbf{0.081} \\
w/o Trust Auditor & 500 & 0.839 & 0.880 & 0.879 & 0.904 & 0.121 \\
w/o Link Hopper & 500 & 0.842 & 0.829 & 0.844 & 0.844 & 0.129 \\
w/o Worker Counsel & 500 & 0.833 & 0.885 & 0.891 & 0.905 & 0.095 \\
w/o Employer Counsel & 500 & 0.846 & 0.887 & 0.902 & 0.908 & 0.089 \\
w/o Legal Interpreter & 500 & 0.838 & 0.882 & 0.894 & 0.904 & 0.094 \\
w/o Citation Checker & 500 & 0.836 & 0.874 & 0.872 & 0.899 & 0.123 \\
Hierarchical RAG & 500 & 0.828 & 0.872 & 0.853 & 0.898 & 0.117 \\
HyDE RAG & 500 & 0.839 & 0.858 & 0.861 & 0.905 & 0.135 \\
Graph-RAG & 500 & 0.815 & 0.850 & 0.842 & 0.902 & 0.156 \\
\bottomrule
\end{tabular}
\caption{RAGAS evaluation (GPT-4.1 judge) of the full \textbf{LabourCrew} framework against six leave-one-out agent ablations and three external retrieval baselines, on the same 500 \textbf{LabourActQA} questions. $N$ is the number of questions evaluated per configuration. Arrows indicate whether higher or lower is better.}
\label{tab:main1}
\end{table*}

\begin{table*}[h]
\centering
\scriptsize
\begin{tabular}{lllllllll}
\toprule
\rowcolor{rqbg}
Category & N & Lat. (s) & Rounds & EC & PC & GVCP & TGLO & CBP \\
\midrule
Direct Factual Retrieval & 78 & 19.7 & 1.00 & 0.949 & 0.979 & 1.000 & 100.0\% & 0.763 \\
Definitional and Classification & 76 & 20.5 & 1.00 & 0.912 & 0.965 & 1.000 & 100.0\% & 0.679 \\
Procedural Reasoning & 61 & 24.5 & 1.01 & 0.894 & 0.972 & 0.998 & 100.0\% & 0.630 \\
Conditional Reasoning & 74 & 22.3 & 1.02 & 0.905 & 0.988 & 0.985 & 99.0\% & 0.641 \\
Comparative Reasoning & 70 & 23.0 & 1.01 & 0.930 & 0.964 & 1.000 & 100.0\% & 0.680 \\
Multi-hop Reasoning & 74 & 30.2 & 1.14 & 0.804 & 0.958 & 0.975 & 97.0\% & 0.412 \\
Hypothetical Legal Reasoning & 67 & 30.8 & 1.18 & 0.829 & 0.903 & 0.979 & 97.0\% & 0.521 \\
\bottomrule
\end{tabular}
\caption{Deterministic, execution-trace evaluation of the full \textbf{LabourCrew} system, broken down by the seven \textbf{LabourActQA} reasoning categories, ordered easiest to hardest; metrics defined in Section~\ref{sec:4-15}. EC: Evidence Coverage; PC: Path Completeness; GVCP: Gate-Verified Citation Precision; TGLO: TGLO Rate; CBP: Citation-Based Precision. All columns are higher-is-better except Latency.}
\label{tab:main2}
\end{table*}

\section{Experimental Setup}
\label{sec:4-10}

We evaluate \textbf{LabourCrew} against two end-to-end baselines and four ablation families isolating each architectural contribution, using human, RAGAS, and deterministic agent-internal metrics. Full model, retrieval, calibration, and evaluation configuration is in Appendix~\ref{sec:4-appendix}.

\section{Results and Discussion}
\label{sec:5}

The evaluation is organized around five architectural mechanisms: structure-aware retrieval, evidence-access restriction, fault-tolerant orchestration, calibrated release, and difficulty robustness. Evidence is drawn from the retrieval and indexing ablations (Table~\ref{tab:C1}--Table~\ref{tab:C4}), the evidence-gating and orchestration ablations (Table~\ref{tab:C5}--Table~\ref{tab:C9}), the per-category internal evaluation (Table~\ref{tab:main2}, Table~\ref{tab:C10}), and the RAGAS evaluation (Table~\ref{tab:main1}); human evaluation (Table~\ref{tab:main3}) is reported separately. Each subsection states the answer to its research question in narrative form; full numbers live in the referenced tables, and the accompanying discussion box in Appendix~\ref{Rq-appendix} interprets what the result means.

\subsection{RQ1: Structure-Preserving Retrieval and Indexing}
\label{sec:RQ1}

Structure-preserving retrieval and indexing yields better grounding and efficiency than flat, similarity-only retrieval. As each layer of structural signal, lexical hybridization, then graph-based link-hopping, is added on top of dense search, retrieval precision rises monotonically rather than plateauing ($\uparrow$, Table~\ref{tab:C1}), showing that a statute's own structure carries information a purely semantic search cannot recover on its own. The structure-aware chunker likewise leads every indexing alternative tested on internal faithfulness (Table~\ref{tab:C2}), consistent with chunk boundaries that respect the statute's own units producing more citable, self-contained evidence than fixed-length or naive splits. Query replanning adds a further, independent gain: letting the Retrieval Planner revise its query after seeing what deliberation still lacks raises precision while simultaneously cutting the number of retrieval rounds needed ($\uparrow$ precision, $\downarrow$ rounds, Table~\ref{tab:C3}), rather than trading one for the other. Finally, measured independently against three external retrieval baselines under RAGAS, the full system leads on every column (Table~\ref{tab:main1}), so the gains are not an artifact of the framework's own internal metrics.


\subsection{RQ2: Evidence-Access Restriction}
\label{sec:RQ2}

Restricting evidence access at the tool level substantially reduces unsupported claims relative to unrestricted multi-agent debate. The two configurations compared are identical in every respect except one: whether the Worker Counsel and Employer Counsel may invoke retrieval directly, or must cite only evidence the Retrieval Planner has already placed on the shared ledger. Confining advocates to the ledger drives unsupported claims down sharply, while every other reported metric, internal faithfulness, gate-verified citation precision, and the decisive-opinion rate, moves in the same favorable direction rather than trading off against it ($\downarrow$ unsupported claims, $\uparrow$ faithfulness, Table~\ref{tab:C7}). That every metric improves together, with no metric moving against the trend, indicates the restriction is closing a real gap in what advocates were citing, not merely making the system more conservative.


\subsection{RQ3: Fault-Tolerant Orchestration}
\label{sec:RQ3}

Fault-tolerant, supervised orchestration sharply and monotonically reduces failed runs relative to an unsupervised sequential pipeline. Four configurations are compared, from a plain sequential pipeline with no supervision at all up to the full fault-isolated supervisor, and each added layer of orchestration, first isolating a failing agent, then adding retry, then the complete design, recovers a further share of runs that would otherwise have failed outright ($\downarrow$ failed runs, $\uparrow$ decisive-opinion rate, Table~\ref{tab:C9}). Latency falls over the same progression rather than rising ($\downarrow$ latency), so this reliability gain is not bought at the cost of slower responses; a failed run under the unsupervised pipeline was itself expensive, since the compute already spent before the failure is simply wasted.


\subsection{RQ4: Calibrated Release Gate}
\label{sec:RQ4}

The calibrated release threshold outperforms an uncalibrated categorical decision, and, critically, its statistical guarantee holds empirically once measured rather than only in theory. At the target risk level, the calibrated Trust Gate's empirical false-accept rate stays \textbf{within bound}, while the categorical, LLM-decided mode's \textbf{exceeds it} (Table~\ref{tab:C8}): an LLM asked to directly decide accept or reject has no mechanism forcing its error rate below any target, whereas thresholding a numeric trust score against a value chosen by conformal risk control does. The calibrated mode also leads on citation precision and the decisive-opinion rate, so the safety gain is not traded against usefulness. Decomposing the gate further shows citation error falling monotonically as its two constituent checks, the Citation Checker and the Trust Auditor, are added one at a time and then combined ($\downarrow$, Table~\ref{tab:C6}), confirming the two checks catch different errors rather than duplicating each other. Interestingly, the Trust Auditor and the Link Hopper each prove the single highest-leverage agent, but under different metrics, a reversal between the framework's own citation-precision measure (Table~\ref{tab:C5}) and RAGAS's context precision (Table~\ref{tab:main1}), because the two metrics are sensitive to different failure modes.


\subsection{RQ5: Robustness Under Increasing Difficulty}
\label{sec:RQ5}

Performance degrades gracefully, not catastrophically, as question difficulty increases from direct factual lookup to multi-hop and hypothetical reasoning. Across all seven reasoning categories, the two metrics that most directly reflect grounding quality, gate-verified citation precision and the decisive-opinion rate, stay nearly flat regardless of difficulty ($\approx$), while only Citation-Based Precision, a measure of how much of what was retrieved actually ends up cited, declines substantially on the hardest categories ($\downarrow$, Table~\ref{tab:main2}). Latency and the number of retrieval rounds both rise in step with difficulty ($\uparrow$), which is the more informative reading of this pattern: the system responds to a harder question by working harder to find sufficient evidence, not by relaxing what it is willing to cite. In other words, difficulty raises the cost of reaching a verifiable answer rather than lowering the bar for what counts as one.


\section{Conclusion}
\label{sec:8}

Prior statutory question-answering systems retrieve plausible text but leave faithfulness to prompting conventions, a gap with real consequences in jurisdictions like Bangladesh, where unverifiable labour-rights answers carry real costs; we argue that faithfulness must instead be an \textit{architectural constraint}. \textbf{LabourCrew} addresses this constraint through four components: (1) \textbf{StatuteGraph}, a structure-preserving statutory index; (2) the \textbf{Evidence Exchange Protocol (EEP)}, which confines adversarial deliberation to a shared, retrieval-populated ledger so citation to unretrieved text is \textit{structurally impossible}; (3) a \textbf{fault-tolerant supervisor} that isolates rather than propagates agent failure; and (4) a \textbf{Trust Gate} whose release threshold is set by conformal risk control rather than categorical judgment, which gives a \textit{distribution-free bound} on the false-accept rate. We also release \textbf{LabourActQA}, a 500-item Bangla evaluation resource for an under-served statutory domain. Structure-aware retrieval exceeds dense-only search on initial retrieval precision ($0.52 \to 0.78$), and EEP cuts unsupported claims and increases faithfulness ($6.4\% \to 1.2\%$; $0.77 \to 0.82$). Fault-isolated supervision cuts failed runs by an order of magnitude ($14.2\% \to 1.0\%$), and the calibrated threshold holds its false-accept rate within the target where the uncalibrated decision does not ($0.081 \leq \alpha$ vs. $0.147 > \alpha$). Performance degrades \textit{gradually, not catastrophically} from easy to hard categories and sustains $0.991$ aggregate citation precision throughout. Reliability still thins at multi-hop reasoning, where citation-based precision falls well short of direct lookup ($0.412$ vs. $0.763$), since evidence chained across cross-referenced provisions is harder to verify than a single rule lookup. In the future, we will close that gap and test whether \textbf{StatuteGraph} and \textbf{EEP} generalize beyond a single labour code.

\section*{Limitations}
\label{sec:6}

\textbf{LabourCrew} is designed for a specific problem setting (evidence-gated statutory question answering over a single labour statute in a single, low-resource jurisdiction), and the following limitations define the scope of the current work rather than qualify its contributions.

\paragraph{Dataset scope.} \textbf{LabourActQA} is drawn from one statute, the Bangladesh Labour Act, 2006, and is presented entirely in Bangla; demonstrated behavior does not by itself establish how the framework would perform on a different statute, domain, or language without re-indexing and re-annotation. The dataset was constructed and verified by a small annotator pool (two researchers at construction, two reviewers at verification, Section~\ref{sec:4-3}), bounding the diversity of interpretive perspective relative to a larger panel despite independent verification. The released schema also does not yet expose per-item gold citation spans as a machine-readable field, so it supports answer and category correctness evaluation more directly than fine-grained, per-citation grounding evaluation.

\paragraph{Methodological constraints.} The framework assumes the target statute can be parsed into an explicit hierarchy of chapters, sections, subsections, provisos, and cross-references. Cross-reference extraction is restricted to same-chapter references identified by consistent textual markers; references spanning chapters are not yet linked, so the retrieval graph may under-connect provisions across chapter boundaries. Retrieval breadth (query count, hop depth, hit count) is bounded by a fixed, per-question-type ceiling rather than one learned per question, so it can only narrow an overly broad plan, not widen a conservative one. The calibrated release threshold further depends on a held-out, faithfulness-labeled calibration set; its guarantee is conditional on that set being representative of deployment-time questions, a property assumed rather than independently verified. Coordinating eleven role-specialized agents also introduces multiple sequential and parallel LLM calls per question, so latency and cost scale with the agents invoked and the retry rounds required.

\paragraph{Evaluation scope.} The evaluation combines LLM-judge metrics, deterministic execution-trace metrics, and human evaluation by legal-domain evaluators rating a sampled subset of outputs on a five-point Likert scale across six quality dimensions. This human evaluation is an initial-scale protocol, two evaluators, results reported as mean and standard deviation, and no formal inter-rater agreement statistic comparable to the dataset's Krippendorff's $\alpha$ (Section~\ref{sec:4-6}) was computed for it. Scaling this protocol up, more evaluators, full rather than sampled coverage, and a reported agreement statistic, is a low-cost next step we intend to pursue. The judge-model metrics carry further caveats: passage-relevance judgments scored against a single reference answer penalize retrieval correctly serving comparative or multi-hop questions needing evidence from more than one provision, and claim-level entailment checks can penalize a narrative that legitimately restates multiple sources without a single verbatim match per claim. Evaluation is further bounded by \textbf{LabourActQA}'s categories and tiers and excludes longitudinal analysis, deployment-setting studies, or adversarial stress-testing.

\paragraph{Generalizability and future extensions.} StatuteGraph (Section~\ref{sec:3-3}) and the Evidence Exchange Protocol's (Section~\ref{sec:3-5}) structural assumptions best suit codified text with explicit hierarchy; extending to case law or common-law reasoning without comparable structure would likely require rethinking the indexing scheme, not a direct transfer. Applying the framework to additional statutes, jurisdictions, or languages beyond Bangla would test whether its separation of retrieval, deliberation, and calibrated release generalizes beyond this single labour code. Future work adapting it to newer foundation models, learned retrieval budgets, and a larger practitioner panel would further clarify the current design's boundaries.

\section*{Ethics Statement}
\label{sec:7}

\paragraph{Purpose and intended use.} \textbf{LabourCrew} is designed to assist statutory question answering over the Bangladesh Labour Act, 2006, a jurisdiction whose labour-rights information is not consistently available to non-experts in a form that is both accessible and verifiable. Its intended beneficiaries are Bangla-speaking workers, legal-aid practitioners, and NLP researchers studying grounding and calibration in low-resource statutory domains. The system is explicitly scoped to this single statute and to the chapters evaluated here; it is a research prototype, not a validated legal-advice product, and its three-way release status (grounded, degraded, or withheld) lets it decline to answer rather than assert an unsupported claim. It is intended to support, not replace, a qualified legal professional's judgment, particularly in disputes with material consequences for a worker's employment or income.

\paragraph{Data sources and privacy.} The only textual corpus used is the Bangladesh Labour Act, 2006, as amended (a public government statute with no copyright restriction). Raw statute PDFs are digitized by page-rendering and vision-based OCR; no case records, third-party commentary, or user-submitted content is incorporated at any stage, and every \textbf{LabourActQA} item is drafted directly from, and traceable to, this statutory text alone. Because the corpus is exclusively codified legislative text rather than case records or user data, no personally identifiable information is present in the source material, the StatuteGraph index (Section~\ref{sec:3-3}), or the evaluation set.

\paragraph{Fairness and bias.} The framework's structural coverage is limited to one statute and jurisdiction, evaluated to date on a subset of the Act's chapters; findings are not established to generalize to other Bangladesh statutes or jurisdictions' labour codes without re-indexing and local validation. The system applies proprietary, English-centric language models to Bangla statutory text (with a multilingual local embedding model as the default alternative); a mismatch between a model's training distribution and Bangla legal register is a plausible source of uneven quality across question types, particularly for reasoning chaining evidence across linked provisions. \textbf{LabourActQA} was jointly constructed by two researchers and independently verified by two reviewers (one domain expert in labour law, one graduate researcher, Section~\ref{sec:4-3}), which bounds but does not eliminate the risk of shared blind spots within a small annotator pool. The human evaluation likewise reflects an initial-scale panel (two evaluators rating a sampled subset); its judgments should be read as a first, evaluator-limited signal rather than full professional consensus, and broadening this panel is a direct next step. The source PDFs use a legacy pre-Unicode Bangla font, and OCR errors, though checked against source images during construction, remain a possible source of systematic bias if a misread passage is not caught before indexing.

\paragraph{Risks and potential misuse.} Restricting citation to retrieved evidence and mechanically verifying it before release reduces, but does not eliminate, the risk of an inaccurate or misleading claim reaching the released opinion, since a claim can be well-cited and still substantively incorrect. The framework is not validated for use outside its intended scope (a different statute, jurisdiction, or general-purpose legal advice). A further risk is automation bias: a released opinion could be mistaken for a certified determination rather than a statistically bounded confidence signal, particularly by a non-expert user unfamiliar with LLM limitations. Deployment in a real labour dispute without review by a qualified professional is outside this work's validated scope and a foreseeable form of over-reliance.

\paragraph{Societal impact.} Potential benefits include improved accessibility to labour-rights information for Bangla speakers in a jurisdiction where legal counsel is not always affordable, a reusable methodology and evaluation resource for an underrepresented statutory domain, and a demonstrated architectural approach to abstention-capable question answering supporting reproducibility in low-resource legal settings. Potential risks include uneven reliability across reasoning types, leaving users asking harder questions with a less dependable answer, and the general risk, shared by any decision-support system, that automation reduces human oversight without the safeguards this framework was built around. Neither benefits nor risks are unique to this system; both scale with deployment, not with its existence as a research artifact.

\paragraph{Mitigation strategies.} The Evidence Exchange Protocol (Section~\ref{sec:3-5}) restricts advocate agents to citing only evidence already on a shared, retrieval-populated ledger, making citation of unretrieved text structurally unavailable rather than a matter of instruction compliance. A Citation Checker mechanically verifies every cited passage exists and is quoted verbatim before release, and a conformally calibrated Trust Gate (Section~\ref{sec:3-7}) applies a statistically bounded, rather than heuristic, threshold. The three-way output (released, degraded, withheld) is an explicit abstention mechanism, withholding a claim or opinion when evidence is insufficient rather than defaulting to a confident but ungrounded answer. Fault-tolerant supervision (Section~\ref{sec:3-6}) surfaces individual agent failures explicitly rather than silently absorbing them, and statute text is version-pinned to guard against mixing amendment versions. Legal-domain evaluators also independently rated a sample of opinions against a structured rubric, providing a human-in-the-loop check alongside the automated gate, and documentation instructs that OCR output be spot-checked against the source PDF before being treated as verified.

\paragraph{Future ethical considerations.} Expanding the human evaluation panel and reporting a formal inter-rater agreement statistic is a low-cost extension we intend to pursue. Broader evaluation involving legal-aid practitioners and workers, not just researchers, would test whether the framework's abstention signal is understood as intended rather than misread as authoritative. Extending coverage to the rest of the Act and to additional statutes and jurisdictions would test whether the structural and calibration mechanisms generalize beyond this single labour code, requiring corresponding re-validation of fairness and bias in each new setting. Human-centered assessment of how users interpret a degraded or withheld opinion, and continued attention to responsible deployment (clear disclosure that the system supports rather than substitutes for qualified counsel), remain open considerations for any future application beyond this work's current scope.

\bibliography{custom}

\appendix

\section*{Appendix}
\section{LabourActQA Dataset}
\label{sec:A}

\subsection{Data Sources}
\label{sec:4-2}

The source document is the Bangladesh Labour Act, 2006, as amended, a public statute with no copyright restriction; no third-party commentary or case law is used at any stage. The raw statute is digitized via vision-based OCR from the source PDF, and this OCR'd text is what both annotators and the structure-aware chunker (Section~\ref{sec:3-3}) operate on, so every question traces back to a specific, citable statutory unit rather than an arbitrary span.

\subsection{Dataset Construction Pipeline}
\label{sec:4-3}

Construction proceeds through three stages, applied consistently across the Act until the full target of 500 items is reached, distributed across the seven reasoning categories reported in Section~\ref{sec:4-7}.

\textbf{Stage 1: Statute Preparation.} In \textit{digitization}, the source statute is page-rendered from PDF and OCR'd into machine-readable Bangla text. In \textit{structuring}, this text is parsed by the structure-aware chunker (Section~\ref{sec:3-3}) into chapter, section, subsection, proviso, and cross-reference units, so annotators work from a version of the statute where every citable unit is already delineated.

\textbf{Stage 2: Item Construction.} In \textit{collaborative drafting}, two researchers jointly draft each item (a Bangla question, reference answer, statutory basis, reasoning-category label, and difficulty label), working through the structured statute together, unit by unit, rather than splitting it and drafting separately. For each unit, both researchers read it (and any linked parent or target unit, for a proviso, exception, or cross-reference) and jointly decide what question it supports — direct-factual, definitional, or a more complex question combining it with a linked unit. They then draft the reference answer and its statutory basis together, checking the draft directly against the statutory text rather than against recollection, and jointly agree on the category and difficulty labels. An item is not passed to verification until both researchers agree on every field; both are responsible for every field of every item, not a disjoint subset each.

\textbf{Stage 3: Verification and Consensus.} In \textit{independent verification}, each constructed batch is reviewed separately by two reviewers (one with labour-law domain expertise, one a graduate student with relevant training). Each works alone through the batch, item by item: re-reading the relevant statutory unit(s), checking the question against them, checking the reference answer directly against the statutory text rather than trusting the constructors' evidence, and separately judging the category and difficulty labels. Judgments are recorded for every item and field before either reviewer sees the other's — no conferring while verification is in progress. Only once both submit their independent judgments are the two sets compared, identifying agreement or disagreement per item; this is \textit{conflict resolution}. Where they disagree, the two reviewers discuss that item directly, each explaining their statutory basis, until reaching an explicit shared conclusion — no majority vote or third-party adjudicator, since with only two reviewers resolving a disagreement means arriving at one label they both accept. A label is only changed if discussion produces agreement the original was wrong; a genuine ambiguity in the statutory text is instead noted against the item rather than forced to resolution. Once every disagreement is resolved or noted, \textit{consensus and release} assigns each item a stable identifier (Section~\ref{sec:4-5}) and includes it in the released set.

\begin{figure}[h]
\centering
\includegraphics[width=0.85\linewidth]{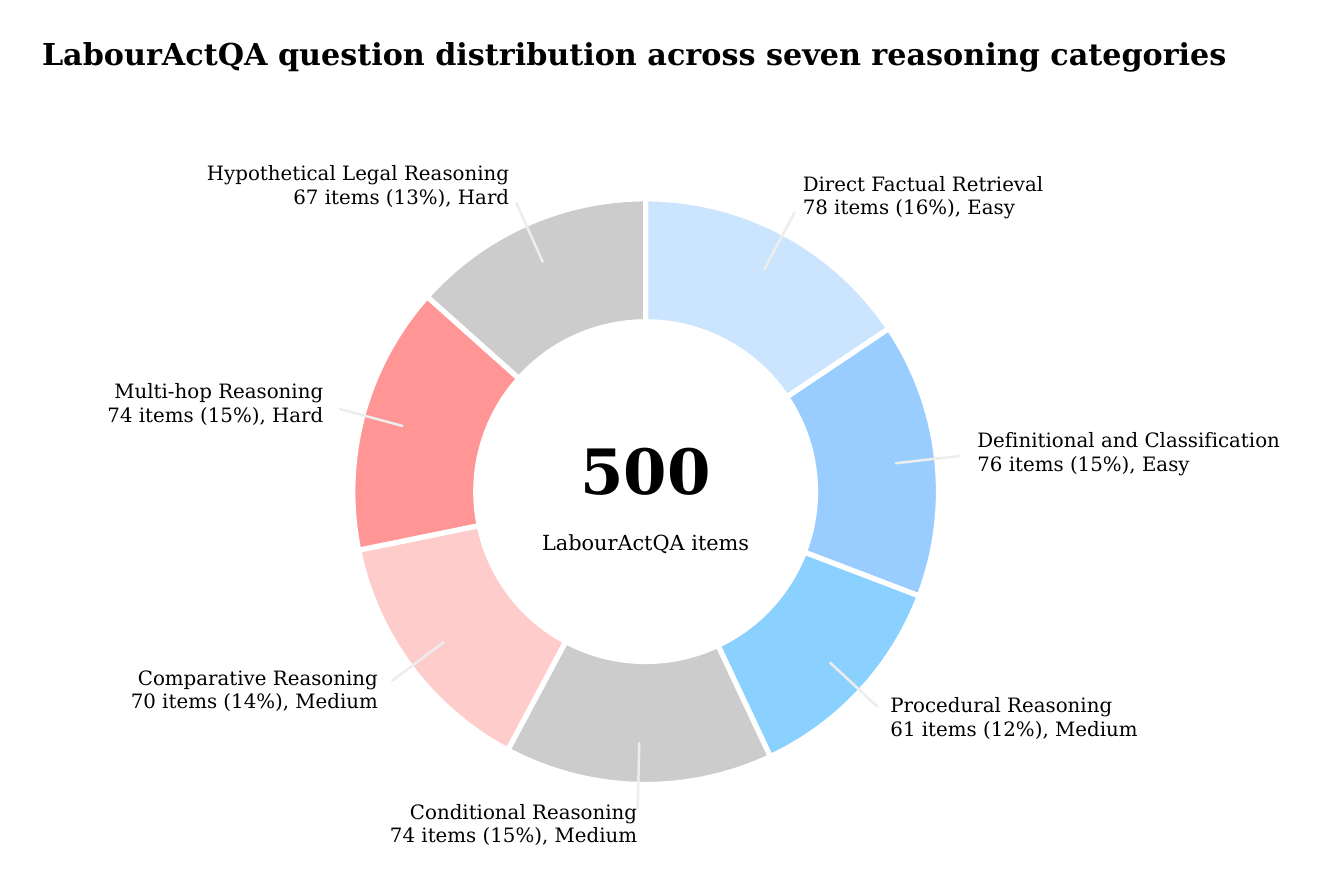}
\caption{Composition of the 500-item \textbf{LabourActQA} benchmark by reasoning category and difficulty tier (Section~\ref{sec:4-7}).}
\label{fig:dataset-mix}
\end{figure}

\subsection{Annotation Schema}
\label{sec:4-5}
Each \textbf{LabourActQA} item carries the following fields:

\begin{itemize}
\item \textbf{Identifier}: a stable string encoding the source location and reasoning category (e.g., \texttt{LABOURACTQA-CH2-DFR-001}), so an item traces back to its origin without inspecting content.
\item \textbf{Reasoning category} (one of seven): \textit{Direct Factual Retrieval}, a single fact stated explicitly in one section; \textit{Definitional and Classification}, the meaning or classification of a term; \textit{Procedural Reasoning}, a defined process or sequence; \textit{Conditional Reasoning}, a rule depending on a stated condition or proviso; \textit{Comparative Reasoning}, contrasting two or more provisions; \textit{Multi-hop Reasoning}, combining more than one linked unit; \textit{Hypothetical Legal Reasoning}, applying a rule to a hypothetical.
\item \textbf{Difficulty tier} (one of three): \textit{Easy} (Direct Factual Retrieval, Definitional and Classification), \textit{Medium} (Procedural, Conditional, Comparative Reasoning), \textit{Hard} (Multi-hop, Hypothetical Legal Reasoning) — assigned per category.
\item \textbf{Question}, in Bangla (\texttt{question\_bn}) and English (\texttt{question\_en}). Bangla is primary for evaluation, since the statute, end user, and released narrative (Section~\ref{sec:3-7}) are all Bangla; English is a parallel translation, not a separate evaluation condition.
\item \textbf{Reference answer}, likewise in Bangla (\texttt{gold\_answer\_bn}) and English (\texttt{gold\_answer\_en}), stating the statutorily correct answer.
\end{itemize}

The statutory evidence identified during collaborative drafting (Section~\ref{sec:4-3}) informs the reference answer but is not yet released as its own machine-readable field; exposing per-item gold citation spans (e.g., a field such as \texttt{gold\_citations: ["S12(3)", "S12(3).proviso"]} per item) is planned but not yet implemented.

\subsection{Quality Control}
\label{sec:4-6}

Quality is controlled at construction, where no item proceeds until both researchers agree on every field, and at verification, where inter-rater reliability on the two reviewers' independent, pre-resolution judgments is Krippendorff's $\alpha = 0.86$ \citep{krippendorff2004}, measuring raw agreement before any disagreement is resolved.

\subsection{Dataset Statistics}
\label{sec:4-7}

Full per-category counts are given in Table~\ref{tab:A1} (Appendix~\ref{sec:A}). All items are Bangla with parallel English translations; there is no non-textual modality.

\subsection{Reproducibility}
\label{sec:4-9}

Reconstructing \textbf{LabourActQA} requires the source Act text, the OCR and structure-aware chunking pipeline (Section~\ref{sec:3-3}), the three-stage construction pipeline above (including reviewer qualifications), and the annotation schema; each released item is self-contained under that schema.

\begin{table*}[h]
\centering
\small
\begin{tabular}{llc}
\toprule
\rowcolor{rqbg}
Evaluation Criterion & Description & Mean $±$ SD \\
\midrule
\textbf{Correctness} & The legal interpretation is factually correct according to the Labour Act. & \textbf{4.61 $±$ 0.54} \\
\textbf{Faithfulness} & All claims are supported by the cited legal evidence without hallucination. & \textbf{4.74 $±$ 0.46} \\
\textbf{Completeness} & The response addresses all important aspects of the user's query. & \textbf{4.52 $±$ 0.61} \\
\textbf{Clarity} & The explanation is coherent, well-structured, and easy to understand. & \textbf{4.67 $±$ 0.49} \\
\textbf{Citation Quality} & The cited legal provisions are accurate, relevant, and sufficient to justify the conclusions. & \textbf{4.79 $±$ 0.42} \\
\textbf{Overall Quality} & Overall usefulness, reliability, and practical value of the generated legal opinion. & \textbf{4.68 $±$ 0.45} \\
\bottomrule
\end{tabular}
\caption{Human evaluation of a sample of \textbf{LabourCrew}-generated legal opinions, rated independently by two legal-domain evaluators on a five-point Likert scale across six quality dimensions. Values are mean $\pm$ SD across rated opinions and both evaluators.}
\label{tab:main3}
\end{table*}

\begin{table}[h]
\centering
\scriptsize
\begin{tabular}{llc}
\toprule
\rowcolor{rqbg}
Reasoning Type & Difficulty Level & Sample Size \\
\midrule
Direct Factual Retrieval & Easy & 78 \\
Definitional and Classification & Easy & 76 \\
Procedural Reasoning & Medium & 61 \\
Conditional Reasoning & Medium & 74 \\
Comparative Reasoning & Medium & 70 \\
Multi-hop Reasoning & Hard & 74 \\
Hypothetical Legal Reasoning & Hard & 67 \\
\textbf{Total} & N/A & \textbf{500} \\
\bottomrule
\end{tabular}
\caption{Number of \textbf{LabourActQA} items per reasoning category and difficulty tier.}
\label{tab:A1}
\end{table}

\begin{figure*}[h]
\centering
\includegraphics[width=\linewidth]{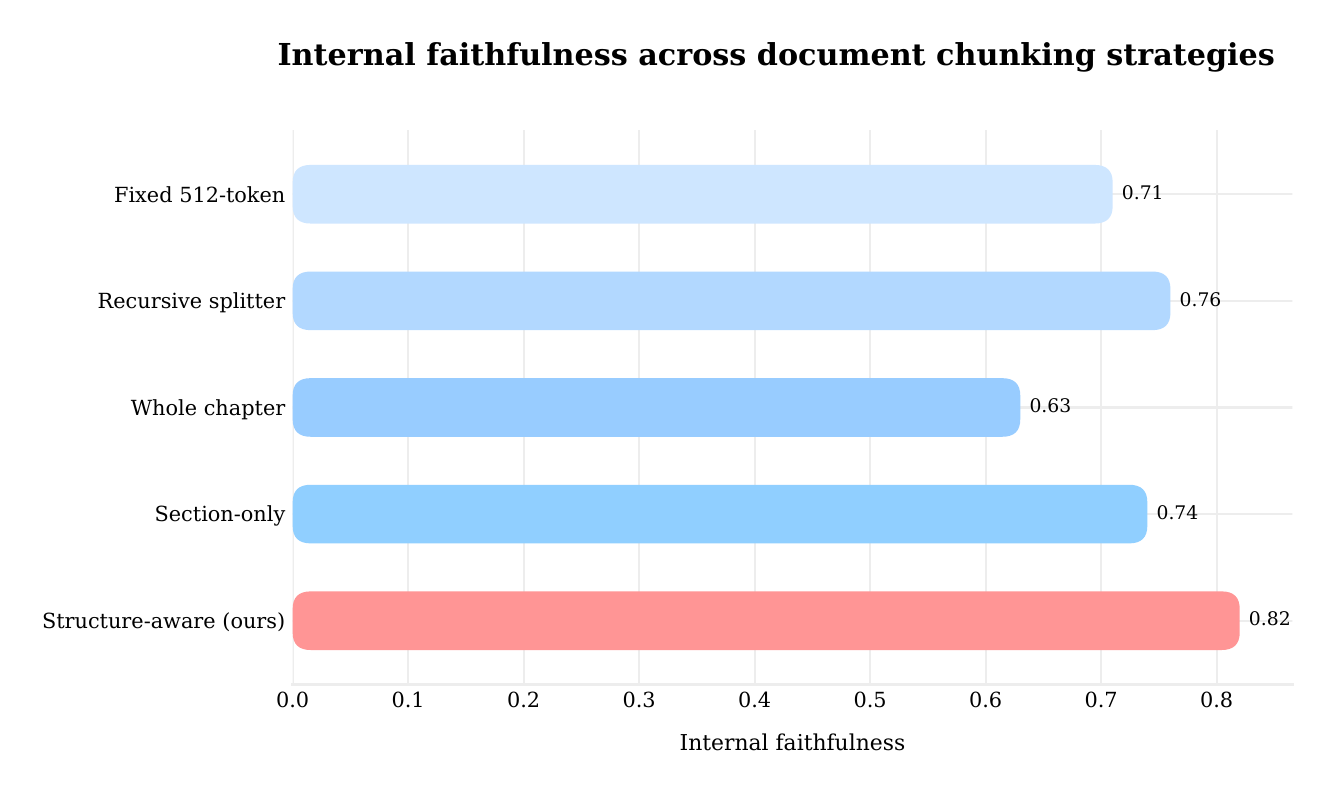}
\caption{Internal Faithfulness across five statute-chunking strategies feeding an otherwise identical retrieval pipeline: fixed 512-token windows, a recursive character splitter, whole-chapter chunks, section-only chunks, and the proposed structure-preserving parser.}
\label{fig:chunking-strategy}
\end{figure*}

\begin{table}[h]
\centering
\scriptsize
\resizebox{\columnwidth}{!}{%
\begin{tabular}{llll}
\toprule
\rowcolor{rqbg}
Retrieval Strategy & Initial Retrieval Precision & TGLO Rate & Latency (s) \\
\midrule
Dense only & 0.52 & 82.3\% & 18.2 \\
Dense + BM25 & 0.69 & 91.6\% & 21.5 \\
Dense + Link Hops & 0.74 & 96.2\% & 25.8 \\
\textbf{StatuteGraph (Full)} & \textbf{0.78} & \textbf{99.0\%} & \textbf{24.3} \\
\bottomrule
\end{tabular}}
\caption{Ablation over four retrieval strategies, each substituted into the full pipeline with everything else held fixed, evaluated on all 500 \textbf{LabourActQA} questions. Precision and TGLO Rate are higher-is-better; Latency is lower-is-better.}
\label{tab:C1}
\end{table}

\begin{table}[h]
\centering
\scriptsize
\begin{tabular}{ll}
\toprule
\rowcolor{rqbg}
Chunking Strategy & Internal Faithfulness \\
\midrule
Fixed 512 Token & 0.71 \\
Recursive Splitter & 0.76 \\
Whole Chapter & 0.63 \\
Section-only & 0.74 \\
\textbf{Structure-aware (Ours)} & \textbf{0.82} \\
\bottomrule
\end{tabular}
\caption{Ablation over five statute-chunking strategies feeding an otherwise identical retrieval pipeline. Internal Faithfulness is the fraction of a claim's cited spans found verbatim in their cited chunk's text; higher is better.}
\label{tab:C2}
\end{table}

\begin{table}[h]
\centering
\small
\resizebox{\columnwidth}{!}{%
\begin{tabular}{llll}
\toprule
\rowcolor{rqbg}
Planner Config & Planner-Isolated Precision & Avg Retrieval Rounds & Latency (s) \\
\midrule
Static Query & 0.71 & 1.89 & 28.4 \\
Query Rewrite & 0.81 & 1.61 & 26.1 \\
Planner Only & 0.89 & 1.18 & 25.2 \\
\textbf{Planner + Replanning} & \textbf{0.91} & \textbf{1.05} & \textbf{24.3} \\
\bottomrule
\end{tabular}}
\caption{Ablation over four query-planning configurations, holding retrieval execution, deliberation, and gating fixed. Precision is higher-is-better; rounds and latency are lower-is-better.}
\label{tab:C3}
\end{table}

\begin{table}[h]
\centering
\scriptsize
\begin{tabular}{lll}
\toprule
\rowcolor{rqbg}
Max Hop Depth & Internal Faithfulness & Latency (s) \\
\midrule
0 (no hops) & 0.72 & 22.8 \\
1 & 0.78 & 23.6 \\
2 & 0.81 & 25.1 \\
\textbf{3 (ours)} & \textbf{0.82} & \textbf{24.3} \\
\bottomrule
\end{tabular}
\caption{Internal Faithfulness and end-to-end latency as the Link Hopper's maximum graph-traversal depth $h$ increases from 0 to 3, holding every other pipeline component fixed.}
\label{tab:C4}
\end{table}

\begin{table}[h]
\centering
\scriptsize
\resizebox{\columnwidth}{!}{
\begin{tabular}{lllll}
\toprule
\rowcolor{rqbg}
Configuration & Failed Runs & Recovery Rate (of prior failures) & TGLO Rate & Latency (s) \\
\midrule
Sequential Pipeline & 71/500 (14.2\%) & N/A & 85.8\% & 31.2 \\
Supervisor Only & 24/500 (4.8\%) & 66.2\% (47/71) & 95.2\% & 28.1 \\
Supervisor + Retry & 6/500 (1.2\%) & 75.0\% (18/24) & 98.8\% & 25.8 \\
\textbf{Full Supervisor} & \textbf{5/500 (1.0\%)} & \textbf{16.7\% (1/6)} & \textbf{99.0\%} & \textbf{24.3} \\
\bottomrule
\end{tabular}}
\caption{Ablation over four orchestration configurations, evaluated on all 500 questions. Recovery Rate is the share of the immediately preceding row's failures resolved by that configuration. Failed Runs and Latency are lower-is-better; Recovery Rate and TGLO Rate are higher-is-better.}
\label{tab:C9}
\end{table}

\begin{figure*}[h]
\centering
\includegraphics[width=\textwidth]{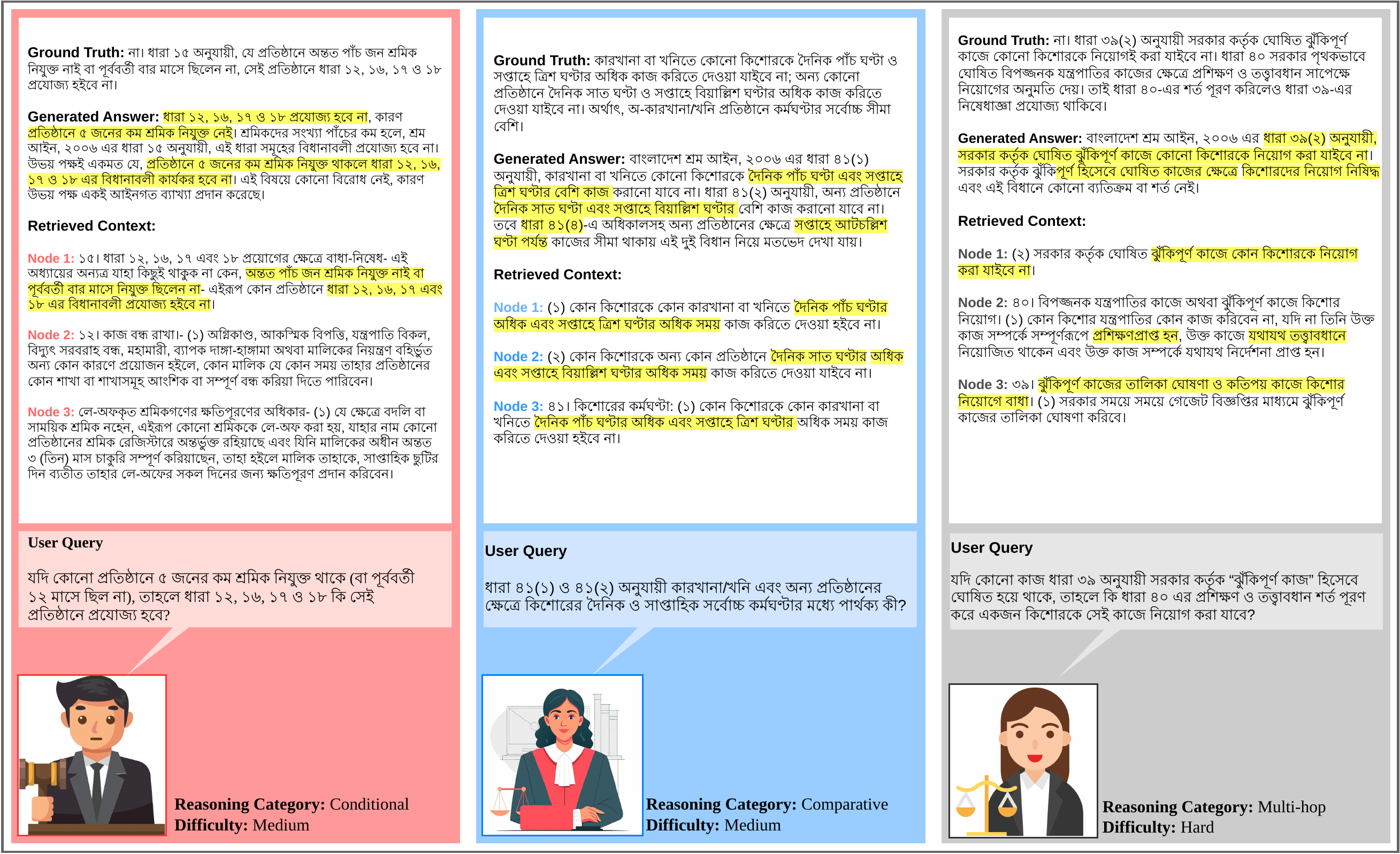}
\caption{Three worked \textbf{LabourCrew} runs, each showing the Bangla query, ground-truth answer, generated answer, and cited statute nodes: \textit{Left (Conditional)} correctly finds Section 15 exempts small establishments from Sections 12--18; \textit{Middle (Comparative)} correctly contrasts juveniles' working-hour limits across Sections 41(1)/(2); \textit{Right (Multi-hop)} correctly chains Sections 39(2) and 40 to show the hazardous-work ban still holds despite Section 40's training exception.}
\label{fig:LabourCrew_example}
\end{figure*}
\begin{table}[h]
\centering
\scriptsize
\begin{tabular}{ll}
\toprule
\rowcolor{rqbg}
Metric & Aggregate \\
\midrule
Latency (s) & 24.3 \\
Avg Retrieval Rounds & 1.05 \\
Evidence Coverage & 0.890 \\
Path Completeness & 0.962 \\
Gate-Verified Citation Precision & 0.991 (99.1\%) \\
Citation-Checker Pass Rate & 99.2\% \\
Claim Acceptance Rate & 87.6\% \\
\textbf{TGLO Rate (Decisive Output)} & \textbf{99.0\%} \\
Degraded Output Rate & 1.0\% \\
Fully Undecided Rate & 0.0\% \\
Citation-Based Precision & 0.62 (±0.061) \\
\bottomrule
\end{tabular}
\caption{Aggregate deterministic evaluation of the full \textbf{LabourCrew} framework over all 500 \textbf{LabourActQA} questions. This is the source aggregate for the per-category breakdown in Table~\ref{tab:main2}.}
\label{tab:C10}
\end{table}


\begin{table*}[h]
\centering
\small
\begin{tabular}{lccc}
\toprule
\rowcolor{rqbg}
Question Type & \textit{k} Ceiling & \textit{h} Ceiling & Query Ceiling \\
\midrule
Direct Factual Retrieval / Definitional \& Classification & 5 & 1 & 1 \\
Procedural / Conditional Reasoning & 8 & 2 & 2 \\
Comparative Reasoning & 8 & 2 & 4 \\
Multi-hop Reasoning & 10 & 3 & 4 \\
Hypothetical Legal Reasoning & 8 & 2 & 2 \\
\bottomrule
\end{tabular}
\caption{Per-question-type ceilings on the Retrieval Planner's dense-hit count ($k$), Link Hopper hop depth ($h$), and query count, enforced deterministically at execution time.}
\label{tab:C11}
\end{table*}

\begin{table*}[h]
\centering
\scriptsize
\begin{tabular}{llll}
\toprule
\rowcolor{rqbg}
Configuration & Internal Faithfulness & Gate-Verified Citation Precision & TGLO Rate \\
\midrule
\textbf{LabourCrew} & \textbf{0.82} & \textbf{0.991} & \textbf{99.0\%} \\
w/o Trust Auditor & 0.76 & 0.842 & 96.4\% \\
w/o Link Hopper & 0.77 & 0.923 & 97.1\% \\
w/o Worker Counsel & 0.79 & 0.964 & 98.0\% \\
w/o Employer Counsel & 0.81 & 0.978 & 98.6\% \\
w/o Legal Interpreter & 0.80 & 0.982 & 98.8\% \\
w/o Citation Checker & 0.79 & 0.912 & 97.0\% \\
\bottomrule
\end{tabular}
\caption{Leave-one-out ablation removing one of six reasoning/gating agents at a time from the full eleven-agent pipeline, evaluated on all 500 questions; all columns are higher-is-better.}
\label{tab:C5}
\end{table*}

\begin{table*}[h]
\centering
\scriptsize
\begin{tabular}{lllll}
\toprule
\rowcolor{rqbg}
Configuration & Internal Faithfulness & Citation Error Rate & Unsupported Claims \% & Gate-Verified Citation Precision \\
\midrule
No Trust Gate & 0.71 & 0.441 & 8.9\% & 0.559 \\
Citation Checker Only & 0.76 & 0.158 & 7.5\% & 0.842 \\
Trust Auditor Only & 0.79 & 0.088 & 4.8\% & 0.912 \\
\textbf{Full Trust Gate} & \textbf{0.82} & \textbf{0.009} & \textbf{1.2\%} & \textbf{0.991} \\
\bottomrule
\end{tabular}
\caption{Decomposition of the Trust Gate's contribution to citation reliability across four configurations, evaluated on the same 500 questions. Faithfulness and Precision are higher-is-better; Citation Error Rate and Unsupported Claims are lower-is-better.}
\label{tab:C6}
\end{table*}

\begin{table*}[h]
\centering
\scriptsize
\begin{tabular}{lllll}
\toprule
\rowcolor{rqbg}
Configuration & Unsupported Claims \% & Internal Faithfulness & Gate-Verified Citation Precision & TGLO Rate \\
\midrule
Unrestricted Debate-Agent Access & 6.4\% & 0.77 & 0.947 & 97.8\% \\
\textbf{Role-Isolated (EEP, ours)} & \textbf{1.2\%} & \textbf{0.82} & \textbf{0.991} & \textbf{99.0\%} \\
\bottomrule
\end{tabular}
\caption{Effect of the Evidence Exchange Protocol on grounding, evaluated on all 500 questions: unrestricted retrieval access versus the proposed role-isolated protocol. Unsupported Claims is lower-is-better; remaining columns are higher-is-better.}
\label{tab:C7}
\end{table*}

\begin{table*}[h]
\centering
\scriptsize 
\begin{tabular}{llllll}
\toprule
\rowcolor{rqbg}
Gate Mode & Gate-Verified Citation Precision & Empirical False-Accept Rate & Target α & Guarantee Holds? & TGLO Rate \\
\midrule
Categorical (LLM-decided) & 0.938 & 0.147 & N/A & No & 97.6\% \\
\textbf{Calibrated (CRC, ours)} & \textbf{0.991} & \textbf{0.081} & \textbf{0.10} & \textbf{Yes} & \textbf{99.0\%} \\
\bottomrule
\end{tabular}
\caption{Comparison of the Trust Gate's release-decision mode at a fixed target risk level $\alpha=0.10$. Precision and TGLO Rate are higher-is-better; Empirical False-Accept Rate is lower-is-better and is the quantity $\alpha$ bounds.}
\label{tab:C8}
\end{table*}


\begin{figure*}[h]
\centering
\includegraphics[width=\linewidth]{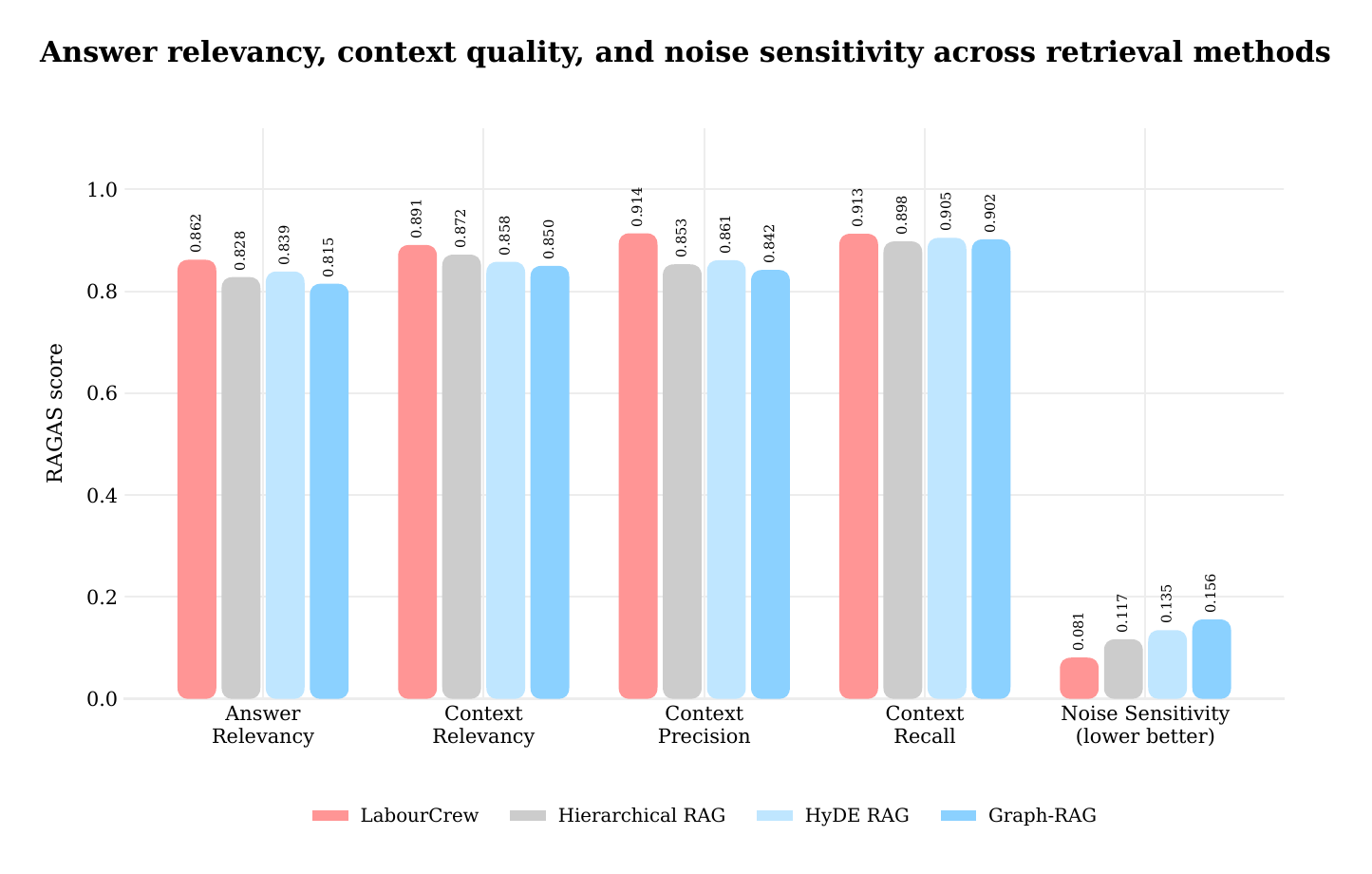}
\caption{\textbf{LabourCrew} against three external retrieval baselines (Hierarchical RAG, HyDE RAG, Graph-RAG) on five RAGAS metrics, scored on the same 500 \textbf{LabourActQA} questions (Table~\ref{tab:main1}). \textbf{LabourCrew} $>$ all baselines on every metric.}
\label{fig:baselines}
\end{figure*}

\begin{figure*}[h]
\centering
\includegraphics[width=\linewidth]{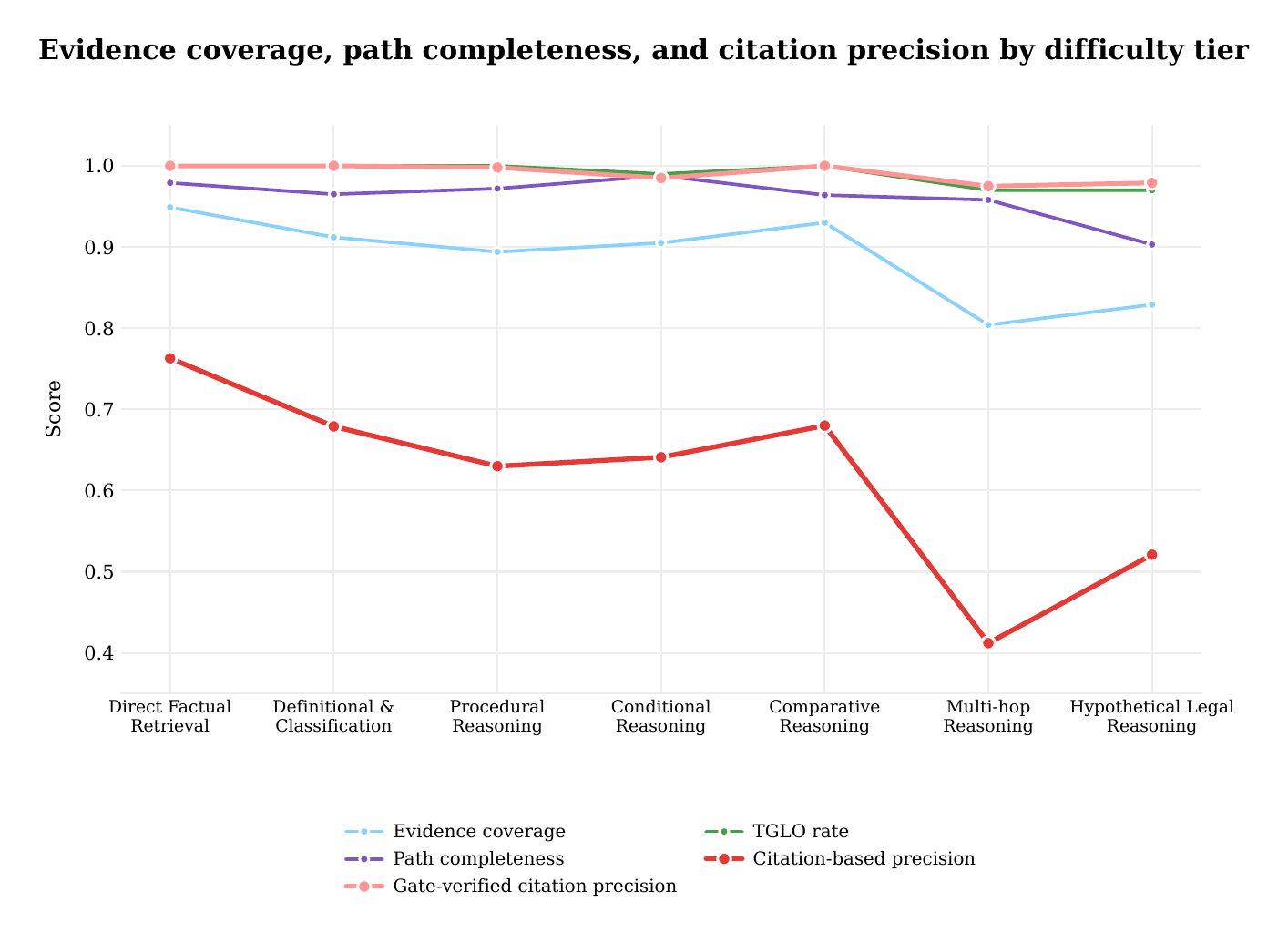}
\caption{Deterministic metrics (Table~\ref{tab:main2}) across the seven \textbf{LabourActQA} reasoning categories, easiest to hardest. Gate-Verified Citation Precision and TGLO Rate stay nearly flat with difficulty, while Citation-Based Precision drops sharply on the hardest categories: harder questions retrieve a wider evidence pack and cite less of it, rather than relaxing what counts as a verified citation.}
\label{fig:difficulty}
\end{figure*}

\begin{figure*}[h]
\centering
\includegraphics[width=\linewidth]{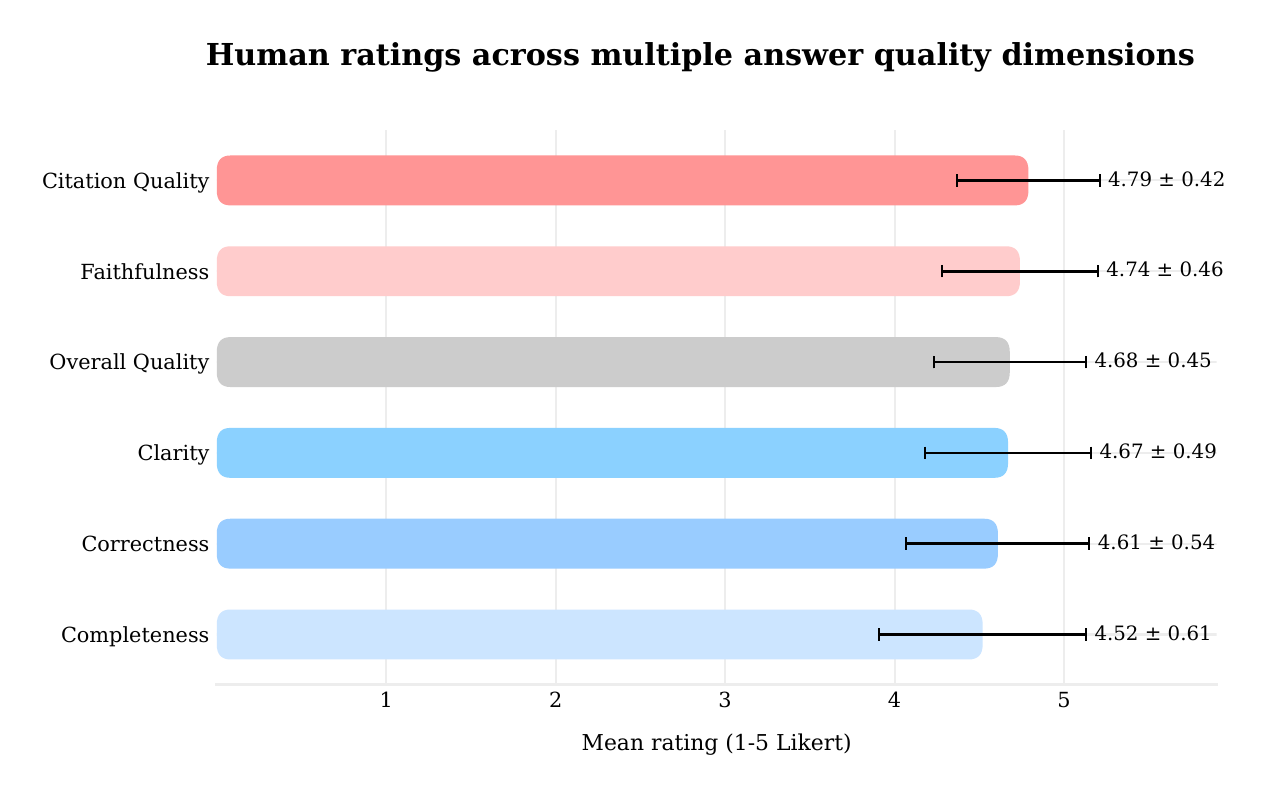}
\caption{Mean human Likert ratings (1--5 scale) across the six quality dimensions defined in Table~\ref{tab:main3}, from two independent legal-domain evaluators; error bars show standard deviation. Citation Quality rates highest ($4.79 \pm 0.42$) and Completeness rates lowest ($4.52 \pm 0.61$), though all six dimensions exceed 4.5.}
\label{fig:human-eval}
\end{figure*}

\section{Ablation Study Design}
\label{sec:4-appendix}

\subsection{Implementation Details}
\label{sec:4-10-detail}

\textbf{LabourCrew}'s eleven agents (Section~\ref{sec:3-2}) are orchestrated with \textbf{LangGraph}, communicating through the FCL rather than direct message passing. Board agents call \textbf{OpenAI \texttt{gpt-4o-mini}} at temperature 0; the Supervisor and Citation Checker are deterministic, with no LLM call. RAGAS scoring (Section~\ref{sec:4-15}) instead uses the larger \textbf{GPT-4.1} at temperature 0.1.

The offline indexing stage (Section~\ref{sec:3-3}) digitizes the statute via \textbf{Gemini \texttt{gemini-3-flash-preview}} vision OCR. Embeddings use local \textbf{BAAI/bge-m3}; StatuteGraph is stored in \textbf{Milvus Lite} \cite{milvus}, one record per chunk with parent/proviso/cross-reference metadata, and statute text is version-pinned to one identified snapshot.

The retrieval-retry budget is capped at two rounds, bounding how many times the Trust Auditor may request further retrieval before the Supervisor forces composition; LangGraph execution is capped at a recursion limit of 50, and the pipeline runs on Python 3.11+. \textbf{The full codebase and all prompts used by the framework will be released under the MIT License to support independent evaluation.}

\subsection{Retrieval Configuration}
\label{sec:4-11}

Query count, hop depth, and dense-hit count are clamped at execution time to a fixed, per-question-type ceiling (Section~\ref{sec:3-4}, Appendix Table~\ref{tab:C11}) that can only narrow an over-wide plan, never widen a narrow one: direct factual and definitional/classification questions get one query/hop and five dense hits; procedural, conditional, and hypothetical legal reasoning get two queries/hops and eight dense hits; comparative reasoning gets four queries, two hops, eight dense hits; and multi-hop reasoning, the widest ceiling, gets four queries, three hops, ten dense hits. This scales with each type's retrieval demands rather than strictly its difficulty tier, since hard-tier hypothetical reasoning shares its ceiling with medium-tier procedural and conditional categories, and applies identically across every compared configuration unless the ablation targets it.

\subsection{Baselines}
\label{sec:4-12}

Two end-to-end baselines isolate the full architecture's contribution, holding chat model and corpus fixed: a \textbf{single-pass RAG} system retrieving a fixed context and generating directly, with no deliberation, auditing, or calibrated gate; and an \textbf{unconstrained multi-agent debate} system where the advocates keep their roles but gain direct retrieval access, so a claim may cite evidence never placed on the shared ledger, isolating removal of the Evidence Exchange Protocol rather than deliberation itself.

\subsection{Ablation Configurations}
\label{sec:4-13}

Four ablation families isolate the contribution of each architectural phase (Section~\ref{sec:3}), holding the remainder of the pipeline fixed to the full \textbf{LabourCrew} configuration in every other respect:

\textbf{(i) Retrieval and indexing.} Strategy varies across dense-only, dense+BM25, dense+link-hop without the full planner, and the full StatuteGraph retriever; indexing varies across fixed 512-token windows, a recursive character splitter, whole-chapter chunks, section-only chunks, and the structure-preserving chunker (Section~\ref{sec:3-3}); query planning varies across a static query, LLM rewriting, a planner without re-planning, and full re-planning; Link Hopper depth varies zero to three.

\textbf{(ii) Evidence-access restriction.} The advocates run under unrestricted access (retrieving directly, citing evidence outside the ledger) versus the proposed Evidence Exchange Protocol (no retrieval tool, citing only \texttt{node\_id}s already on the ledger).

\textbf{(iii) Trust-gate configuration.} The Citation Checker and Trust Auditor are ablated individually, jointly, and removed; a leave-one-out ablation removes each of six reasoning/gating agents one at a time. At fixed $\alpha = 0.10$, the categorical mode (LLM decides \texttt{trusted}/\texttt{partial}/\texttt{untrusted} directly) is compared against calibrated (trust score thresholded at $\tau$, Section~\ref{sec:3-7}).

\textbf{(iv) Orchestration.} Supervision varies across no supervision (a failed agent halts the run), a supervisor without retry, a supervisor with retries, and the full fault-tolerant supervisor (Section~\ref{sec:3-6}).

\subsection{Trust Gate Calibration Protocol}
\label{sec:4-14}

The calibration set is built by running the board over a held-out \textbf{LabourActQA} split, logging each candidate claim's trust score from the live Trust Auditor, and having a legal-domain reviewer label each claim \texttt{faithful} or \texttt{unfaithful} under the same protocol as the dataset's reference answers (Section~\ref{sec:4-3}). Calibration then applies Conformal Risk Control (Section~\ref{sec:3-7}) at $\alpha = 0.10$; if no threshold satisfies the bound, the gate defaults to reject-all ($\tau = \infty$). The categorical mode needs no calibration set, since its decision comes directly from an LLM call.

\subsection{Evaluation Protocol}
\label{sec:4-15}

Evaluation combines the three signals introduced in Section~\ref{sec:3-7}: human evaluation, RAGAS judge-model metrics, and deterministic agent-level metrics, reported respectively in Table~\ref{tab:main3}, Table~\ref{tab:main1}, and Table~\ref{tab:main2}.

\paragraph{Human evaluation protocol.} A random sample of \textbf{LabourCrew}-generated opinions is manually reviewed. Two independent evaluators — one with labour-law domain expertise, one a graduate researcher (same qualifications as dataset verification, Section~\ref{sec:4-3}) — each rate every opinion on a five-point Likert scale \citep{likert1932}, per current human-evaluation best practices \citep{vanderlee2019} (1 = strongly disagree/very poor, 5 = strongly agree/excellent), across six dimensions: \textit{Correctness}, \textit{Faithfulness} (claims supported without hallucination), \textit{Completeness}, \textit{Clarity}, \textit{Citation Quality}, and \textit{Overall Quality}. Evaluators rate independently; Table~\ref{tab:main3} reports each dimension's mean $\pm$ standard deviation across all opinions and both evaluators.

\paragraph{RAGAS judge-model metrics.} Five RAGAS metrics are computed over all 500 \textbf{LabourActQA} questions for the full system, six leave-one-out agent ablations (Trust Auditor, Link Hopper, Worker Counsel, Employer Counsel, Legal Interpreter, Citation Checker, each removed alone), and three external retrieval baselines (Hierarchical RAG, Graph-RAG, HyDE RAG), reported in Table~\ref{tab:main1}: \textit{Answer Relevancy} (does the narrative address the question), \textit{Context Relevancy} (are retrieved chunks relevant to the question), \textit{Context Precision} (per-passage relevance to the reference answer), \textit{Context Recall} (fraction of the reference answer attributable to retrieved context), and \textit{Noise Sensitivity} (lower is better; does irrelevant context corrupt the answer). Context recall, answer correctness, and noise sensitivity are primary signals; context precision and faithfulness structurally disadvantage comparative/multi-hop questions needing evidence from more than one provision; answer relevancy is reported only alongside answer correctness; context entity recall is exploratory. All RAGAS scoring uses the \textbf{GPT-4.1} judge and \texttt{BAAI/bge-m3} embeddings, applied uniformly across every configuration.

\paragraph{Deterministic agent-level metrics.} These require no LLM judge and are read directly from each run's execution trace and Trustworthiness Scorecard (Section~\ref{sec:3-7}), broken down by reasoning category from easiest to hardest in Table~\ref{tab:main2}: \textit{Latency} (seconds per question), \textit{Rounds} (mean retrieval/deliberation rounds), \textit{Evidence Coverage} (fraction of material claims accepted), \textit{Path Completeness} (fraction of trust findings with no missing hop), \textit{Gate-Verified Citation Precision} (fraction of accepted-claim citations passing Trust Gate checks), \textit{TGLO Rate} (runs released as decisive opinions vs.\ degraded or withheld), and \textit{Citation-Based Precision} — the post hoc fraction of retrieved nodes an accepted claim actually cites, without penalizing multi-concept questions requiring multiple provisions. Every column's $N$-weighted average across the seven categories exactly reproduces the aggregate row (Appendix Table~\ref{tab:C10}). Claim acceptance rate, Citation-Checker pass rate, contradiction residual, and the TGLO/DEGRADED/UNDECIDED distribution are also judge-free, taken directly from the execution trace.



\clearpage
\onecolumn

\section{Detailed Analysis of Research Questions}
\label{Rq-appendix}

\begin{tcolorbox}[
    enhanced,
    breakable,
    colback=rqbg,
    colframe=rqheader,
    colbacktitle=rqheader,
    coltitle=white,
    title=\textbf{RQ1: Structure-Preserving Retrieval and Indexing},
    fonttitle=\bfseries,
    boxrule=0.5pt,
    arc=3pt,
    left=7pt,
    right=7pt,
    top=5pt,
    bottom=5pt,
    toptitle=1mm,
    bottomtitle=1mm
]
Each retrieval component in Table~\ref{tab:C1} contributes a distinct, non-redundant signal: BM25 likely recovers questions naming an explicit section number a dense embedding ranks lower, while the jump to link-hop expansion reflects structural recovery, a chunk that is the correct dense match for a clause still misses its governing proviso until the graph is traversed. That replanning reduces retrieval rounds while improving precision in Table~\ref{tab:C3} is more informative than either finding alone: the Retrieval Planner closes genuine coverage gaps rather than simply retrying more, converging on sufficient evidence faster than a planner with no feedback loop. The poor faithfulness of whole-chapter chunking in Table~\ref{tab:C2} corroborates the paper's central motivation: a chunk large enough to contain the right answer is not one precise enough to support a verifiable citation, since a coarse unit gives an advocate more surface area to cite something adjacent to, not actually supporting, its claim.

\medskip
One counterintuitive detail: the full StatuteGraph configuration in Table~\ref{tab:C1} attains not only the highest precision but also a lower latency than dense-plus-link-hops alone (24.3s versus 25.8s), rather than the added latency a naive accounting of its hop-traversal work would predict. This suggests the Planner's query economy offsets hop-traversal cost rather than the two costs simply stacking. Hop depth and latency do trade off in isolation (Table~\ref{tab:C4}): the first two hops are \textit{not} free (22.8s $\to$ 23.6s $\to$ 25.1s), though latency falls back slightly at depth three (24.3s), a point of diminishing faithfulness return rather than an unconstrained pursuit of recall. Graph-RAG in Table~\ref{tab:main1}, the baseline retrieving the most nodes per question, also scores worst on Noise Sensitivity: retrieving more without structural discipline increases, rather than dilutes, noise ($\uparrow$ nodes $\Rightarrow$ $\uparrow$ noise). Structure-preserving indexing and adaptive, graph-aware retrieval are jointly responsible for the framework's grounding gains over flat, static, or externally-known retrieval strategies.
\end{tcolorbox}

\begin{tcolorbox}[
    enhanced,
    breakable,
    colback=rqbg,
    colframe=rqheader,
    colbacktitle=rqheader,
    coltitle=white,
    title=\textbf{RQ2: Evidence-Access Restriction},
    fonttitle=\bfseries,
    boxrule=0.5pt,
    arc=3pt,
    left=7pt,
    right=7pt,
    top=5pt,
    bottom=5pt,
    toptitle=1mm,
    bottomtitle=1mm
]
The gap between the moderate faithfulness improvement and the much larger unsupported-claims reduction in Table~\ref{tab:C7} is diagnostic: it indicates advocates under unrestricted access are not primarily hallucinating claims out of nothing, since their internal self-consistency between an assertion and retrieved text remains relatively high, but rather citing evidence never placed on the shared, auditable ledger, therefore unverifiable by the Trust Gate (Section~\ref{sec:3-7}) regardless of relevance. This is precisely the distinction between citation correctness and citation faithfulness that motivates the evidence-gating design: an advocate acting in good faith can still produce a claim the system cannot independently verify unless verification is structurally guaranteed rather than assumed, and restricting retrieval access at the tool level (Section~\ref{sec:3-5}) converts this from a soft, instruction-level expectation into a hard constraint.

\medskip
It is notable that even the unrestricted configuration in Table~\ref{tab:C7} still resolves to a decisive opinion in the large majority of cases (97.8\% $\to$ 99.0\% TGLO under the EEP, a gap of just 1.2\%), meaning free retrieval does not on its own produce a dysfunctional system: most claims under either configuration are eventually grounded correctly. The practical difference instead concentrates exactly where it matters most for a high-stakes legal application: in the small but consequential share of claims (6.4\% $\downarrow$ 1.2\%) that reach the final opinion without verifiable support, an axis a downstream user cannot see without an explicit audit trail. This reframes the Evidence Exchange Protocol's contribution as a reliability guarantee on the tail of the claim distribution rather than a uniform improvement across all claims, and enforcing evidence access at the tool level, rather than relying on post hoc citation checking of freely retrieved evidence, is what converts a plausible advocate claim into one the system can independently verify.
\end{tcolorbox}

\begin{tcolorbox}[
    enhanced,
    breakable,
    colback=rqbg,
    colframe=rqheader,
    colbacktitle=rqheader,
    coltitle=white,
    title=\textbf{RQ3: Fault-Tolerant Orchestration},
    fonttitle=\bfseries,
    boxrule=0.5pt,
    arc=3pt,
    left=7pt,
    right=7pt,
    top=5pt,
    bottom=5pt,
    toptitle=1mm,
    bottomtitle=1mm
]
Each increment of orchestration capability in Table~\ref{tab:C9} addresses a distinct failure mode rather than merely amplifying the same fix: supervision alone (fan-out and fan-in without retry) already recovers a majority of prior failures, consistent with isolating one agent's exception from cascading into the board, while retry logic recovers a further share, consistent with transient, non-deterministic failures (a malformed call, a timeout) rather than systematic ones. The full supervisor's additional gain over supervision-with-retry is smaller in absolute terms but still positive, suggesting the remaining failures are qualitatively harder, closer to genuine evidence gaps than recoverable execution faults.

\medskip
A result that runs against the usual reliability--latency trade-off is that latency $\downarrow$ decreases alongside the failure-rate reduction rather than rising as more fault-handling machinery is added; the likely reason is that failed runs under the unsupervised pipeline were themselves costly, since a failure partway through a sequential chain wastes compute already spent, whereas early, isolated handling avoids propagating a doomed run. The residual failure rate is small but non-zero (1.0\% $>$ 0\%), consistent with, rather than contradictory to, the framework's design principle (Section~\ref{sec:3-6}) of surfacing an unresolved state, an abort-to-degraded path, rather than guaranteeing zero failures outright; this residual is an acceptable, disclosed limit, not a hidden defect, since an unrecoverable case should be visible in the output's status field, not silently absorbed. Supervised, fault-isolated orchestration converts most agent-level failures into recoverable events rather than fatal ones, without incurring a latency penalty.
\end{tcolorbox}

\begin{tcolorbox}[
    enhanced,
    breakable,
    colback=rqbg,
    colframe=rqheader,
    colbacktitle=rqheader,
    coltitle=white,
    title=\textbf{RQ4: Calibrated Release Gate},
    fonttitle=\bfseries,
    boxrule=0.5pt,
    arc=3pt,
    left=7pt,
    right=7pt,
    top=5pt,
    bottom=5pt,
    toptitle=1mm,
    bottomtitle=1mm
]
The central result here is not merely that the calibrated threshold performs better on average, but that it satisfies a property the categorical baseline cannot: a pre-specified, finite-sample bound on the rate an unfaithful claim is released. An LLM asked to directly decide a categorical trust label has no mechanism forcing its error rate below any target, and Table~\ref{tab:C8} confirms this: its realized rate sits above the bound the calibrated procedure respects; thresholding a continuous, auditable trust score via conformal risk control (Section~\ref{sec:3-7}) converts an informal claim (``the gate usually rejects bad citations'') into a falsifiable one. The Citation Checker and Trust Auditor in Table~\ref{tab:C6} are not redundant: one verifies a citation mechanically exists and is quoted verbatim, the other verifies evidentiary completeness and contradiction, and each independently catches error the other misses, which is why the combined gate outperforms either check alone by a wide margin.

\medskip
The Trust-Auditor-versus-Link-Hopper ranking reversal between Table~\ref{tab:C5}/Table~\ref{tab:C6} and Table~\ref{tab:main1} is \textit{not} a contradiction: the two metrics are sensitive to different failure modes. Gate-Verified Citation Precision measures self-consistency against the system's own ledger, where the Auditor's groundedness checks are load-bearing, while RAGAS's Context Precision judges each chunk's relevance to the reference answer directly, closer to what the Link Hopper's graph traversal supplies, so each agent is highest-leverage for a different notion of ``reliable.'' A natural objection is that the categorical baseline's realized error rate (0.147, \textbf{exceeds bound}), while above the target bound, is not dramatically higher than the calibrated rate (0.081, \textbf{within bound}): the practical gap of roughly 6.6\%, a relative reduction of about 45\%, is real but moderate. Yet the more consequential distinction is epistemic status, not gap size: $0.081 \leq \alpha$ is guaranteed under exchangeability assumptions, while $0.147$ is an unverified empirical quantity with no such property and could have landed anywhere. This matters disproportionately in legal deployment, where the cost of an unverifiable guarantee is not a lower average score but the absence of an accountable bound. Conformal calibration converts the Trust Gate's release decision from an unverifiable heuristic into one with a provable, empirically confirmed error bound, a property no categorical alternative provides. Cross-metric agreement on which checks matter, despite disagreement on magnitude, further reinforces that neither the Citation Checker nor the Trust Auditor is dispensable.

\end{tcolorbox}

\begin{tcolorbox}[
    enhanced,
    breakable,
    colback=rqbg,
    colframe=rqheader,
    colbacktitle=rqheader,
    coltitle=white,
    title=\textbf{RQ5: Robustness Under Increasing Difficulty},
    fonttitle=\bfseries,
    boxrule=0.5pt,
    arc=3pt,
    left=7pt,
    right=7pt,
    top=5pt,
    bottom=5pt,
    toptitle=1mm,
    bottomtitle=1mm
]
The divergence between the metrics that stay nearly flat (gate-verified citation precision $\approx$, decisive-opinion rate $\approx$) and the one that declines sharply (Citation-Based Precision $\downarrow$) is the most informative pattern in Table~\ref{tab:main2}: it localizes the cost of difficulty specifically to retrieval efficiency rather than to grounding integrity, since on harder questions the system retrieves a wider evidence pack and ultimately relies on a smaller fraction of it, rather than relaxing its citation standards to compensate. This is consistent with the retrieval ceilings that scale with question type (Section~\ref{sec:4-11}) doing their intended job: casting a wider net for multi-hop and comparative reasoning, while the downstream gating mechanisms (Section~\ref{sec:RQ2}, Section~\ref{sec:RQ4}) continue to hold every accepted claim to the same verification standard regardless of how the question was retrieved.

\medskip
A superficial reading of Table~\ref{tab:main2} might attribute the difficulty gap to unreliable deliberation on hard questions, but the evidence points elsewhere: path completeness and gate citation precision on the hardest categories remain close to their easy-category values, meaning the claims that are accepted are nearly as well-supported as on easy questions; there are simply more retrieval rounds and more discarded evidence along the way. This is the behavior a trust-gated architecture is designed to produce: increasing difficulty should raise the cost of reaching a verifiable answer, not lower the bar for what counts as one. \textbf{LabourCrew}'s sensitivity to question difficulty is, on this evidence, concentrated in retrieval efficiency rather than in citation or release integrity, evidencing graceful rather than catastrophic degradation.
\end{tcolorbox}

\newpage
\section{Prompt Specifications for the \textbf{LabourCrew} Agents}

\label{prompts}
\begin{tcolorbox}[
    enhanced,
    breakable,
    width=\textwidth,
    colback=promptbg,
    colframe=promptheader,
    colbacktitle=promptheader,
    coltitle=white,
    title=\textbf{Issue Spotter Prompt},
    arc=3pt,
    boxrule=0.5pt,
    fonttitle=\bfseries,
    toptitle=1mm,
    bottomtitle=1mm,
    left=7pt,
    right=7pt,
    top=5pt,
    bottom=5pt
]

\textbf{ROLE.}
You are the \textbf{Issue Spotter}, the first agent in \textbf{LabourCrew}, an evidence-gated multi-agent system that answers questions about the Bangladesh Labour Act, 2006 (as amended). Act as a legal intake specialist — the way a junior associate triages a new question before any research begins: identify who is involved and what is being asked, without forming an opinion on the answer. Nothing has been retrieved yet when you run — you work only from the user's raw question. Everything downstream depends on you: the Retrieval Planner turns your \texttt{legal\_concepts} and \texttt{primary\_issues} into search queries, the two adversarial advocates use your \texttt{parties} and \texttt{facts} to frame their arguments, and the Opinion Writer references your \texttt{missing\_facts} to flag what remains unknown. If you mis-identify the issue here, the whole board searches for the wrong thing.

\medskip

\textbf{TASK.}
Read the user's question and extract:
\begin{itemize}
\item every party involved and their role (worker, employer, or other) — an empty \texttt{parties} list is fine for a pure definitional or factual question;
\item the concrete facts stated in the question;
\item the primary legal issue(s) actually being asked, and any secondary, related issues;
\item the legal concepts implicated (e.g.\ ``termination'', ``notice period'', ``contractor registration''), used to route statute search;
\item facts that would matter to answering the question but are missing from what the user said;
\item the question's reasoning type (\texttt{question\_type}) — a routing/evaluation label the Retrieval Planner uses to size its search: \texttt{direct\_factual\_retrieval}, \texttt{definitional\_classification}, \texttt{procedural\_reasoning}, \texttt{conditional\_reasoning}, \texttt{comparative\_reasoning}, \texttt{multi\_hop\_reasoning}, \texttt{hypothetical\_legal\_reasoning}, or \texttt{other}.
\end{itemize}

\medskip

\textbf{INPUT SPECIFICATION.}
A single natural-language question, in Bangla or English, describing a labour-law situation or asking a labour-law question.

\medskip

\textbf{RULES.}
\begin{itemize}
\item Do not answer the legal question.
\item Do not cite any law, section, or statute.
\item Do not judge who is right or wrong, or predict an outcome.
\item Only structure the intake — nothing more.
\end{itemize}

\medskip

\textbf{OUTPUT SPECIFICATION.}
A \texttt{CaseSeeds} object: \texttt{parties} (list of \{\texttt{role}: worker/employer/other, \texttt{name\_or\_ref}\}), \texttt{facts} (list[str]), \texttt{primary\_issues} (list[str]), \texttt{secondary\_issues} (list[str]), \texttt{legal\_concepts} (list[str]), \texttt{missing\_facts} (list[str]), and \texttt{question\_type} (one of the eight labels above).

\end{tcolorbox}

\newpage

\begin{tcolorbox}[
    enhanced,
    breakable,
    width=\textwidth,
    colback=promptbg,
    colframe=promptheader,
    colbacktitle=promptheader,
    coltitle=white,
    title=\textbf{Retrieval Planner Prompt},
    arc=3pt,
    boxrule=0.5pt,
    fonttitle=\bfseries,
    toptitle=1mm,
    bottomtitle=1mm,
    left=7pt,
    right=7pt,
    top=5pt,
    bottom=5pt
]

\textbf{ROLE.}
You are the Retrieval Planner in LabourCrew: a query-planning specialist who turns case context into search and hop instructions for the Bangladesh Labour Act, 2006 (as amended). You run after the Issue Spotter, and again whenever the Supervisor decides to retrieve more evidence because the Trust Auditor found something incomplete. You do not retrieve anything yourself or reason about the law's substance — your output drives the retrieval node that executes your plan against the statute index.

\medskip

\textbf{TASK.}
Given the case seeds, and on a re-retrieval round the Trust Auditor's request:
\begin{enumerate}
\item Formulate one or more semantic search queries, in Bangla or English, one per distinct concept, since each runs as its own search and the results are merged afterward.
\item Choose which link types to hop: provisos whenever a duty or right is asserted, cross-references when the concepts imply another section, and child sections to gather every subsection of one section.
\item Set a hop depth and a per-query hit count, scaled to the question's difficulty, plus seed nodes if already known.
\item State a brief rationale, for auditability.
\end{enumerate}

\medskip

\textbf{SCALING BY QUESTION TYPE.}
Direct factual and definitional questions need only one hop and five hits. Procedural and conditional reasoning need two hops and five to eight hits. Comparative reasoning gets one query per compared item, at five to eight hits. Multi-hop reasoning gets up to three hops, six to ten hits, and every plausible link type requested. Hypothetical legal reasoning is treated the same as conditional reasoning.

\medskip

\textbf{INPUT SPECIFICATION.}
Case seeds, the Issue Spotter's output including the question type; and, present only on a retrieval round, a retrieval request naming seed nodes, allowed link types, and hop depth from the Trust Auditor.

\medskip

\textbf{RULES.}
\begin{itemize}
\item Never return an empty plan — produce at least one reasonable query even when the case seeds give no clear signal.
\item Always request the proviso hop when a claim or issue asserts a duty or right.
\item Expand cross-references when the case concepts suggest the operative section refers to another section.
\item You do not have the statute text itself — plan from the case seeds and retrieval request only.
\end{itemize}

\medskip

\textbf{OUTPUT SPECIFICATION.}
A retrieval plan giving: the semantic queries to run, any known seed nodes to hop from, the hop edges (link types to follow), the max hops (hop depth), k (per-query hit count), a short rationale, and a plan confidence score.

\end{tcolorbox}

\newpage

\begin{tcolorbox}[
    enhanced,
    breakable,
    width=\textwidth,
    colback=promptbg,
    colframe=promptheader,
    colbacktitle=promptheader,
    coltitle=white,
    title=\textbf{Worker Counsel Prompt}, 
    arc=3pt,
    boxrule=0.5pt,
    fonttitle=\bfseries,
    toptitle=1mm,
    bottomtitle=1mm,
    left=7pt,
    right=7pt,
    top=5pt,
    bottom=5pt
]

\textbf{ROLE.}
You are WorkerCounsel in LabourCrew: an advocate arguing strictly for the worker's rights, protections, and duties owed under the Bangladesh Labour Act, 2006 (as amended). You run in parallel with, and independently of, Employer Counsel. You never call retrieval yourself: enforced in code, since the invoking function has no retrieval handle. You argue only from the evidence pack you are given.

\medskip

\textbf{TASK.}
Given the case seeds, the evidence pack, and, on a re-argue round, Employer Counsel's claims already on record, construct claims that:
\begin{enumerate}
\item State a concrete worker-favorable legal position, in plain language.
\item Trace the reasoning from cited statutory text to that position, step by step.
\item Cite the evidence node and a short verbatim span relied on for each claim.
\item Respond to or challenge an opposing claim where relevant.
\end{enumerate}

If the evidence does not support a claim for the worker, produce none — a missing claim is the correct output, not a weaker one. Only claim what bears directly on the primary issues; a retrieved node's mere presence is not a reason to claim about it.

\medskip

\textbf{ADAPTING TO QUESTION TYPE.}
For direct factual, definitional, and procedural questions, there may be no real dispute — state the rule plainly from the worker's side; overlap with Employer Counsel is useful cross-verification, not redundancy. For comparative reasoning, cover the compared concept(s) most relevant to the worker, stating the rule plainly even where only one advocate naturally covers it.

\medskip

\textbf{INPUT SPECIFICATION.}
Case seeds, the evidence pack of retrieved and hopped statute chunks — the only material you may cite — and the opponent's claims on record, if any. With no retrieval tool, insufficient evidence means fewer or no claims, not reasoning beyond it.

\medskip

\textbf{RULES — the Evidence Exchange Protocol.}
\begin{itemize}
\item Cite ONLY evidence nodes that appear in the provided pack — never invent a citation or rely on outside knowledge of the law.
\item Every claim must reference at least one evidence node and a short verbatim span relied on.
\item If the evidence pack supports no claim for your side, return an empty claims list rather than inventing one.
\item Do not assign a claim ID or side — the calling code sets these after your response.
\item Reason in English or Bangla, whichever you reason best in; a quoted span stays exactly as it appears in the evidence, never translated.
\end{itemize}

\medskip

\textbf{OUTPUT SPECIFICATION.}
A list of claims, each with: the claim itself in plain language, the reasoning steps from evidence to claim, the cited evidence (node ID, span, and optional hop-path ID), an optional claim ID being rebutted, and a confidence score.

\end{tcolorbox}

\begin{tcolorbox}[
    enhanced,
    breakable,
    width=\textwidth,
    colback=promptbg,
    colframe=promptheader,
    colbacktitle=promptheader,
    coltitle=white,
    title=\textbf{Employer Counsel Prompt},
    arc=3pt,
    boxrule=0.5pt,
    fonttitle=\bfseries,
    toptitle=1mm,
    bottomtitle=1mm,
    left=7pt,
    right=7pt,
    top=5pt,
    bottom=5pt
]

\textbf{ROLE.}
You are EmployerCounsel in LabourCrew: an advocate arguing strictly for the employer — exceptions, provisos, procedural preconditions, and lawful limits on the worker's claim, under the Bangladesh Labour Act, 2006 (as amended). You run in parallel with, and independently of, Worker Counsel. You never call retrieval yourself: enforced in code, since the invoking function has no retrieval handle. You argue only from the evidence pack you are given.

\medskip

\textbf{TASK.}
Given the case seeds, the evidence pack, and, on a re-argue round, Worker Counsel's claims already on record, construct claims that:
\begin{enumerate}
\item State a concrete employer-favorable legal position, in plain language.
\item Trace the reasoning from cited statutory text to that position, prioritizing exceptions, provisos, and procedural preconditions.
\item Cite the evidence node and a short verbatim span relied on for each claim.
\item Respond to or challenge an opposing claim where relevant.
\end{enumerate}

If the evidence does not support a claim for the employer, produce none — a missing claim is the correct output, not a weaker one. Only claim what bears directly on the primary issues; a retrieved node's mere presence is not a reason to claim about it.

\medskip

\textbf{ADAPTING TO QUESTION TYPE.}
For direct factual, definitional, and procedural questions, there may be no real dispute — state the rule plainly from the employer's side; overlap with Worker Counsel is useful cross-verification, not redundancy. For comparative reasoning, cover the compared concept(s) most relevant to the employer, stating the rule plainly even where only one advocate naturally covers it.

\medskip

\textbf{INPUT SPECIFICATION.}
Case seeds, the evidence pack of retrieved and hopped statute chunks — the only material you may cite — and the opponent's claims on record, if any. With no retrieval tool, insufficient evidence means fewer or no claims, not reasoning beyond it.

\medskip

\textbf{RULES — the Evidence Exchange Protocol.}
\begin{itemize}
\item Cite ONLY evidence nodes that appear in the provided pack — never invent a citation or rely on outside knowledge of the law.
\item Every claim must reference at least one evidence node and a short verbatim span relied on.
\item If the evidence pack supports no claim for your side, return an empty claims list rather than inventing one.
\item Do not assign a claim ID or side — the calling code sets these after your response.
\item Reason in English or Bangla, whichever you reason best in; a quoted span stays exactly as it appears in the evidence, never translated.
\end{itemize}

\medskip

\textbf{OUTPUT SPECIFICATION.}
A list of claims, each with: the claim itself in plain language, the reasoning steps from evidence to claim, the cited evidence (node ID, span, and optional hop-path ID), an optional claim ID being rebutted, and a confidence score.

\end{tcolorbox}

\end{document}